\documentclass[10pt,journal,compsoc]{IEEEtran}
\usepackage{graphicx}
\usepackage{amsmath}
\usepackage{amssymb}
\usepackage{booktabs}
\usepackage[switch]{lineno}

\usepackage[dvipsnames]{xcolor}
\usepackage{algorithm}
\usepackage[noend]{algpseudocode}

\usepackage{pgfpages}
\usepackage{pgfplots}
\usepackage{pgfplotstable,}
\pgfplotsset{compat=newest}

\pgfplotsset{
        /pgfplots/ybar legend/.style={
        /pgfplots/legend image code/.code={%
        \draw[##1,/tikz/.cd,bar width=3pt,yshift=-0.2em,bar shift=0pt]
                plot coordinates {(0cm,0.8em)};},
},
}

\usepackage{tikz-dependency}
\usetikzlibrary{patterns}

\newcommand{\graphtrailer}{\textsc{Graph\-Trailer}\xspace}
\newcommand{\textrank}{\textsc{Text\-Rank}\xspace}
\newcommand{\graphtp}{\textsc{Graph\-TP}\xspace}

\usepackage{graphicx}
\usepackage{mathtools}
\usepackage{array}
\newcolumntype{L}[1]{>{\raggedright\let\newline\\\arraybackslash\hspace{0pt}}m{#1}}
\newcolumntype{C}[1]{>{\centering\let\newline\\\arraybackslash\hspace{0pt}}m{#1}}
\newcolumntype{R}[1]{>{\raggedleft\let\newline\\\arraybackslash\hspace{0pt}}m{#1}}
\usepackage{amsfonts}
\usepackage{xspace}

\usepackage[size=scriptsize]{todonotes}
\usepackage{arydshln}
\usepackage{color}
\usepackage{xcolor,soul}

\usepackage[inline]{enumitem}

\definecolor{one}{HTML}{16697a}
\definecolor{two}{HTML}{008891}
\definecolor{three}{HTML}{db6400}

\definecolor{tp1}{HTML}{66CC00}
\definecolor{tp2}{HTML}{009900}
\definecolor{tp3}{HTML}{999900}
\definecolor{tp4}{HTML}{FF0000}
\definecolor{tp5}{HTML}{FF6666}

\ifCLASSOPTIONcompsoc
  \usepackage[nocompress]{cite}
\else
  \usepackage{cite}
\fi
\ifCLASSINFOpdf
\else
\fi

\renewcommand{\paragraph}[1]{\vspace*{1.5ex}\hspace*{-.5cm}\textbf{#1}~}

\begin{document}
%
\title{Finding the Right Moment: Human-assisted Trailer
  Creation via Task Composition}
%
%
%
%

\author{Pinelopi Papalampidi,
        Frank Keller,
        and~Mirella Lapata
\IEEEcompsocitemizethanks{\IEEEcompsocthanksitem Pinelopi Papalampidi, Frank Keller and Mirella Lapata are with the School of Informatics, University of Edinburgh, UK.\protect\\
E-mail: p.papalampidi@sms.ed.ac.uk, \{keller,mlap\}@inf.ed.ac.uk}
\thanks{Manuscript received July 19, 2022.}}

%
%

\markboth{IEEE Transactions on Pattern Analysis and Machine Intelligence}%
{Shell \MakeLowercase{\textit{et al.}}: Bare Demo of IEEEtran.cls for Computer Society Journals}
%



\IEEEtitleabstractindextext{%
\begin{abstract}
  Movie trailers perform multiple functions: they introduce viewers to
  the story, convey the mood and artistic style of the film, and
  encourage audiences to see the movie. These diverse functions make
  trailer creation a challenging endeavor. In this work, we focus on
  finding \textit{trailer moments} in a movie, i.e., shots that could
  be potentially included in a trailer. We decompose this task into
  two subtasks: narrative structure identification and sentiment
  prediction. We model movies as graphs, where nodes are shots and
  edges denote semantic relations between them. We learn these
  relations using \textit{joint contrastive training} which distills
  rich textual information (e.g.,~characters, actions, situations)
  from screenplays. An unsupervised algorithm then traverses the graph
  and selects trailer moments from the movie that human judges prefer
  to ones selected by competitive supervised approaches. A main
  advantage of our algorithm is that it uses interpretable criteria,
  which allows us to deploy it in an interactive tool for trailer
  creation with a human in the loop. Our tool allows users to select
  trailer shots in under 30 minutes that are superior to fully
  automatic methods and comparable to (exclusive) manual selection by
  experts.
\end{abstract}

}

\maketitle

\IEEEdisplaynontitleabstractindextext

%
\IEEEpeerreviewmaketitle

\IEEEraisesectionheading{\section{Introduction}\label{sec:introduction}}

\IEEEPARstart{T}{railers} are short videos used for promoting movies
and are often critical to commercial success. While their core
function is to market the film to a range of audiences, trailers are
also a form of persuasive art and promotional narrative, designed to
make viewers want to see the movie. Even though the making of trailers
is considered an artistic endeavor, the film industry has developed
strategies guiding trailer construction. According to one school of
thought, trailers must exhibit a narrative structure, consisting of
three
acts.\footnote{https://www.studiobinder.com/blog/how-to-make-a-movie-trailer}
The first act establishes the characters and setup of the story, the
second act introduces the main conflict, and the third act raises the
stakes and provides teasers from the ending. Another school of thought
is more concerned with the mood of the trailer as defined by the ups
and downs of the
story.\footnote{https://www.derek-lieu.com/blog/2017/9/10/the-matrix-is-a-trailer-editors-dream}
According to this approach, trailers should have medium intensity at
first in order to captivate viewers, followed by low intensity for
delivering key information about the story, and then progressively
increasing intensity until reaching a climax at the end of the
trailer.

In this work we aim to \emph{automatically} identify moments in a
movie that are suitable for including in a trailer. For this, we need
to perform low-level tasks such as person identification, action
recognition, and sentiment prediction, but also more high-level ones
such as understanding connections between events and their causality,
as well as drawing inferences about the characters and their actions.
Given the complexity of the task, \emph{directly} learning all this
knowledge from movie--trailer pairs would require many thousands of
examples, whose processing and annotation would be a challenge. It is
thus not surprising that previous approaches
\mbox{(\hspace*{-.5ex}\cite{irie2010automatic,smeaton2006automatically,wang2020learning})}
have solely focused on audiovisual features and depend on ill-defined
criteria, such as identifying the ``trailerness'' of a movie shot.

Following previous
work~\mbox{(\hspace*{-.5ex}\cite{irie2010automatic,smeaton2006automatically,wang2020learning})},
we formulate moment identification as the task of selecting
\textit{shots} for presentation in a trailer, under the assumption
that a segmentation of the movie into moments (i.e., shots that
represent events) is available.  Inspired by the creative process of
human trailer editors, we adopt a bottom-up approach, decomposing our
task into two simpler subtasks. The first one is narrative structure
analysis, i.e.,~retrieving the most important events of the movie. A
commonly adopted theory in screenwriting
\mbox{(\hspace*{-.5ex}\cite{cutting2016narrative,Hauge:2017,thompson1999storytelling})}
suggests that there are five types of key events in a movie, known as
turning points (TPs; see their definitions in Figure~\ref{fig:TPs}).
The second subtask is sentiment prediction, which we view as an
approximation of the intensity flow between shots and the emotions
they evoke.

\begin{figure}[t]
\centering
\begin{small}
    \begin{tabular}{@{}p{3cm}@{~~}p{5.6cm}@{}} \hline
 \raisebox{-2.5ex}[0pt]{1. Opportunity} & Introductory event that occurs after presentation of
 setting and      background of main characters. \\ \hline
 \raisebox{-1.5ex}[0pt]{2. Change of Plans} &   Main goal of story is defined; action begins to
increase.\\ \hline 
     \raisebox{-1.5ex}[0pt]{3. Point of No Return} & Event that pushes the characters to
      fully commit to their goal.  \\\hline
 \raisebox{-1.5ex}[0pt]{4. Major Setback} &  Event where everything falls apart,
      temporarily or permanently. \\ \hline
 \raisebox{-1.5ex}[0pt]{5. Climax} &  Final event of the main story,  moment of resolution.  \\ \hline
\end{tabular}
\end{small}
\caption{\label{fig:TPs} Turning points and
  their definitions \cite{Hauge:2017}. 
}
\end{figure}

\begin{figure}[t]
    \centering
\includegraphics[width=\columnwidth]{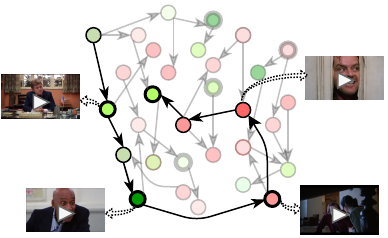}
\caption{\graphtrailer performs walks (bold line) in a movie graph to
  generate proposal trailer sequences. Nodes in the graph are shots
  and edges denote relations between them. Each shot is characterized
  by a sentiment score (green/red shades for positive/negative values)
  and labels describing important events (thick circles).}
    \label{fig:intro}
\end{figure}

We identify trailer moments following an unsupervised graph-based
approach. We model movies as graphs whose nodes are shots and whose
edges denote important semantic connections between shots (see
Figure~\ref{fig:intro}). In addition, nodes bear labels denoting
whether they are key events (i.e.,~TPs) and scores signaling sentiment
intensity (positive or negative). Our algorithm traverses this movie
graph to retrieve sequences of movie shots that can be used in a
trailer. In contrast to prior work, we exploit \textit{all modalities}
(i.e.,~video, audio, text) for identifying trailer moments. More
importantly, our method uses interpretable criteria (i.e.,~key events,
sentiment scores) and therefore can be deployed as part of an
interactive trailer creation tool with a human in the loop.


Both tasks of TP analysis and sentiment prediction stand to benefit
from lower-level understanding of movie content. Indeed, we could
employ off-the-shelf modules for identifying characters and places,
recognizing actions, and localizing semantic units. However, such
approaches substantially increase pre-processing time and memory
requirements during training and inference and suffer from error
propagation. Instead, we propose a contrastive learning regime, where
we take advantage of screenplays as \emph{privileged
  information}~\cite{lopez2015unifying}, i.e., information available
at \emph{training time} only. Screenplays reveal how the movie is
segmented into scenes,
who and where the characters are, when and who they are speaking to,
what they are doing (in a screenplay, ``scene headings'' explain where
the action takes place while ``action lines'' describe what the camera
sees).  Specifically, we build two networks, a \textit{multimodal}
network based on movie videos and an \textit{auxiliary, textual}
network based on screenplays, and train them jointly using contrastive
losses in order to distill information from screenplays to videos.  The
auxiliary network can be additionally pretrained on large collections
of screenplays via self-supervised learning, without processing the
corresponding movies, overcoming data scarcity issues.  Experimental
results show that contrastive training is beneficial for knowledge
distillation, leading to trailers which are judged favorably by
annotators in terms of their content and attractiveness.

Finally, we explore how our algorithm can be used in an interactive
setting where users select shots to be included in trailers from a set
of automatically identified candidates. We provide an interactive tool
for trailer
creation\footnote{https://movie-trailers-beta.herokuapp.com} and
assess its functionality against fully manual and fully automatic
methods. Our study  reveals that interactive selection improves the
quality of the trailers and is comparable to selecting shots via
manual inspection, while reducing the time needed from 2--3~days to
under 30~minutes. Our contributions in this work can be summarized as
follows:
\begin{itemize}
    \item We propose an unsupervised approach for identifying
      shot-level trailer moments in movies which operates over sparse
      graphs and decomposes the task into  \emph{narrative structure}
      analysis  and \emph{sentiment} prediction. 

    \item We propose a \textit{contrastive} training regime for
      \textit{distilling knowledge} from screenplays to movies in lieu
      of collecting and processing full-length movie videos.

    \item We develop an interactive tool for building trailers
      \textit{semi-automatically} and demonstrate they are better than
      fully automatic ones and of similar
      quality to   trailers created exclusively by human experts. 
\end{itemize}

\section{Related Work}

\textbf{Movie understanding} approaches have mainly focused on
isolated video clips, and tasks such as the alignment between movie
scenes and book chapters \cite{tapaswi2015book2movie}, question
answering (\hspace*{-.8ex}\cite{tapaswi2016movieqa,lei2018tvqa}),
video captioning for movie shots \cite{rohrbach2015dataset} or clips
from TV episodes~\cite{lei2020tvr}, and text-to-video retrieval
\mbox{(\hspace*{-.5ex}\cite{bain2020condensed,lei2020tvr,liu2020violin}}). Although
this work exploits multimodal information (i.e.,~mainly video and
language), it does not target full-length movies. Bain et
al.~\cite{bain2020condensed} adopt a more holistic approach and learn
from movie clips, corresponding textual descriptions, and other
metadata, such as genre information and bounding boxes for
characters. Follow-on work~\cite{huang2020movienet} creates a
large-scale dataset based on full-length movies, their trailers,
posters, textual synopses, scripts, action recognition tags, and
character bounding boxes. Unfortunately, this dataset is not publicly
available, limiting further research.  Some recent work
\mbox{(\hspace*{-.5ex}\cite{papalampidi2019movie,papalampidi2020screenplay,papalampidi2020movie,chen2021summscreen})}
attempts to identify narrative structure and summarize entire TV
episodes and movies, but focuses exclusively on the textual modality
(i.e.,~screenplays). Our approach exploits information from multiple
dimensions, i.e.,~video, audio, text, whilst processing full-length
movies.

 \paragraph{Trailer moment identification} exploits superficial
audiovisual features, such as background music or visual changes
between
shots~\mbox{(\hspace*{-.5ex}\cite{irie2010automatic,smeaton2006automatically})}. Other
work creates ``attractive'' trailers with a graph-based model for shot
selection \cite{xu2015trailer} or uses a human in the loop in
conjunction with a model trained on horror movies via audiovisual
sentiment analysis \cite{smith2017harnessing}. The Trailer Moment
Detection Dataset \cite{wang2020learning} consists of full-length
movies paired with official trailers and annotations for key moments,
but is not publicly available and does not include screenplays. Wang
et al.~\cite{wang2020learning} propose a state-of-the-art trailer
generation model that again only focuses on the visual modality. To
the best of our knowledge, we are the first to exploit all input
modalities for identifying trailer moments in full-length movies.

\paragraph{Video highlight detection}
\mbox{(\hspace*{-.5ex}\cite{ye2021temporal,badamdorj2021joint,chen2021pr})},
i.e., identifying frames or video segments that can function as
highlights is also relevant to our work. Most previous approaches
focus on short videos with simple semantics (e.g.,~actions in YouTube
videos), do not exploit the textual modality, and cannot easily
transfer to the task of trailer moment identification. An adjacent
line of work focuses on video editing
(\hspace*{-.8ex}\cite{gu2020sumbot,frey2021automatic,shen2022autotransition,argaw2022anatomy,pardo2022moviecuts}),
in order to create cohesive and visually appealing videos. Although
transitions between shots are stylistically important, our model does
not take this into account when selecting shots for inclusion in the
trailer.  However, it is possible for humans to post-process the
trailer through our interactive tool.

\paragraph{Knowledge distillation}
\mbox{(\hspace*{-.5ex}\cite{ba2014deep,hinton2015distilling})} was
originally proposed for distilling information from a larger teacher
model to a smaller student one. Generalized distillation
\cite{lopez2015unifying} provides a framework for using privileged
information available at training time only.  Most related to our work
is the use of different modalities/views of the same content
\mbox{(\hspace*{-.5ex}\cite{miech2019howto100m,miech2020end})}, such
as transcribed narrations to learn visual representations in
videos. We leverage screenplays as a source of privileged information
and distill knowledge about events, characters, and scenes in a film,
which we exploit for identifying trailer-worthy shots in video.

\section{Problem Formulation} \label{sec:problem_formulation}
We are interested in \textit{automatically} identifying  important
movie content that should be included in a trailer.  We define trailer
moment identification as the task of selecting $L$~shots from a
full-length movie with~$M$ shots ($L \ll M$).  A \emph{shot} is the
continuous sequence of frames between two edits or cuts in a movie.
The selected shots do not constitute the final trailer, several
post-processing steps might be required: trimming them, changing their
order, adding music, voice-over, and other information (e.g.,~release
date), however we leave this to future work.

Movies are typically complex stories, they contain distinct subplots
or events that unfold non-linearly, while redundant events, called
``fillers'' enrich the main story. We therefore cannot assume that
consecutive shots are semantically related. To better explore
relations between events, we represent movies as graphs
\cite{papalampidi2020movie}. Let $\mathcal{G}=(\mathcal{V},
\mathcal{E})$ denote a graph where vertices $\mathcal{V}$ are shots
and edges~$\mathcal{E}$ represent their semantic similarity.  We
further consider the original temporal order of shots in $\mathcal{G}$
by only allowing directed edges from previous to future
shots. $\mathcal{G}$~is described by an upper-triangular transition
matrix~$\mathcal{T}$, which records the probability of transitioning
from shot~$i$ to every future shot~$j$.

We propose an algorithm for traversing~$\mathcal{G}$ and selecting
sequences of shots to be used in a trailer. The algorithm operates
over graph structures exemplified in Figure~\ref{fig:intro}:
$\mathcal{G}$~represents relations between movie shots describing key
events (TPs; thick circles in Figure~\ref{fig:intro}); in addition
each node in the graph has a sentiment score which can be positive or
negative (different shades of green/red depending on the sentiment
intensity in Figure~\ref{fig:intro}).
 In the following, we first describe our algorithm
 (Section~\ref{sec:graph_traversal}) assuming graph~$\mathcal{G}$ is
 given and then discuss how~$\mathcal{G}$ is learned and key events
 are detected \cite{papalampidi2019movie}
 (Section~\ref{sec:tp_identification}). Finally, we explain how
 shot-based sentiment scores are predicted
 (Section~\ref{sec:sentiment_prediction}).

\subsection{\graphtrailer: Movie Graph Traversal} \label{sec:graph_traversal}

\begin{algorithm}[t]
\caption{Graph traversal: Retrieve trailer path}\label{algorithm}
\begin{small}
\textbf{Input:} shot-level graph $\mathcal{G}$, sets of TP shots,
sentiment scores for all shots \\ 
\textbf{Output:} proposal trailer path (Path)
\begin{algorithmic}[1]
\Procedure{\graphtrailer}{}
\State {\small Path $\gets \emptyset$,  budget $\gets L$,  TPs\_id $\gets 0$,  flow $\gets \emptyset$}

\State events $\gets$ {\small$[\mathcal{TP}_1, \mathcal{TP}_2, \mathcal{TP}_3, \mathcal{TP}_4, \mathcal{TP}_5]$} 

\State i $\gets$ sample({\small$\mathcal{TP}_1$}) 

\State add i to Path

\State next\_TP $\gets$ events[TPs\_id] 

\While {budget $> 0$ $\And$ TPs\_id $< 5$}

\State next\_node $\vcentcolon= \operatornamewithlimits{argmax}\limits_{j \in \mathbb{N}_{i}}(s_{ij})$ \Comment{Eq.~\ref{eq:score}}
 
\State add next\_node to Path

\State add sentiment(next\_node) to flow

\State i $\gets$ next\_node

\State budget -= 1
\If {i $\in$ next\_TP $\cup$ $\mathcal{N}_{next\_TP}$}
\State {\small TPs\_id++, \, next\_TP $\gets$ events[TPs\_id]}

\EndIf

\EndWhile

\Return Path

\EndProcedure
\end{algorithmic}
\end{small}
\end{algorithm}

Algorithm~\ref{algorithm} retrieves trailer sequences by performing
random walks in graph~$\mathcal{G}$. We start by selecting a node
identified as a first TP (i.e.,~Opportunity, see
Figure~\ref{fig:TPs}), i.e.,~the first major event that changes the
course of the plot and serves as a catalyst for character development;
we assume that first TPs should be included in a trailer as they
reflect what the movie is about. Note that TPs extend over $C$~shots
and as a result, our algorithm can produce $C$~different paths as
proposal trailers (see line~4 in Algorithm~\ref{algorithm}).  Given
node~$i$, we decide where to go next by considering its $K$ immediate
neighbors~$\mathcal{N}_i$. We select node~$j$ from~$\mathcal{N}_i$ as
the~$k^\mathit{th}$ shot~$n_k$ to add to the path based on the
following criteria: (1)~normalized probability of transition~$e_{ij}$
from~$i$ to~$j$ based on matrix~$\mathcal{T}$ (i.e.,~\textit{semantic
  similarity} between shots), (2)~normalized distance $t_{ij}=|j-i|/M$
between~$i$ and~$j$ (i.e.,~\textit{temporal proximity}),
(3)~normalized shortest path\footnote{Computed using Dijkstra's
\cite{Dijkstra:1959} algorithm.} from node $j$ to next major event
  $d_{j,\mathcal{TP}}$ (i.e.,~\textit{relevance to the storyline}),
  and (4)~difference between sentiment flow~$p_{ij}$ (from~shot $i$
  to~$j$) and
\textit{expected} flow~$f_k$ at the $k^\mathit{th}$ step in the path (see
Figure~\ref{fig:intro} and Appendix, Section~1.3):
\begin{gather}
n_k =
\operatornamewithlimits{argmax}\limits_{j\in\mathcal{N}_i}s_{ij} \label{eq:next_shot} \\
    s_{ij} = \lambda_1e_{ij} - \lambda_2t_{ij} - \lambda_3d_{j,\mathcal{TP}} -\lambda_4|p_{ij}-f_k| \label{eq:score}
\end{gather}
where $\lambda_1, \lambda_2, \lambda_3, \lambda_4$ are hyperparameters
used to combine the different criteria (tuned on the development set
based on gold-standard trailer labels).  Note that these criteria are
interpretable and can be easily altered by a user (e.g.,~by adding new
ones or defining a different flow~$f$). Our approach can also be used
for interactive trailer creation, where a user iteratively decides
which shot to include in the trailer from a limited set of options
(see Section~\ref{sec:discussion} for details).

We select $L$ shots in total (depending on a target trailer length)
and retrieve a proposal trailer sequence as depicted in
Figure~\ref{fig:intro} (bold line). At each step, we keep track of the
sentiment flow created and the TPs identified thus far (lines~10
and~13--14 in Algorithm~\ref{algorithm}, respectively). A TP event is
 selected for presentation in the trailer if a shot or its
immediate neighbors have been added to the path.

\subsection{Graph Construction and TP Identification} \label{sec:tp_identification} 

\begin{figure*}
    \centering
    \includegraphics[width=0.8\textwidth]{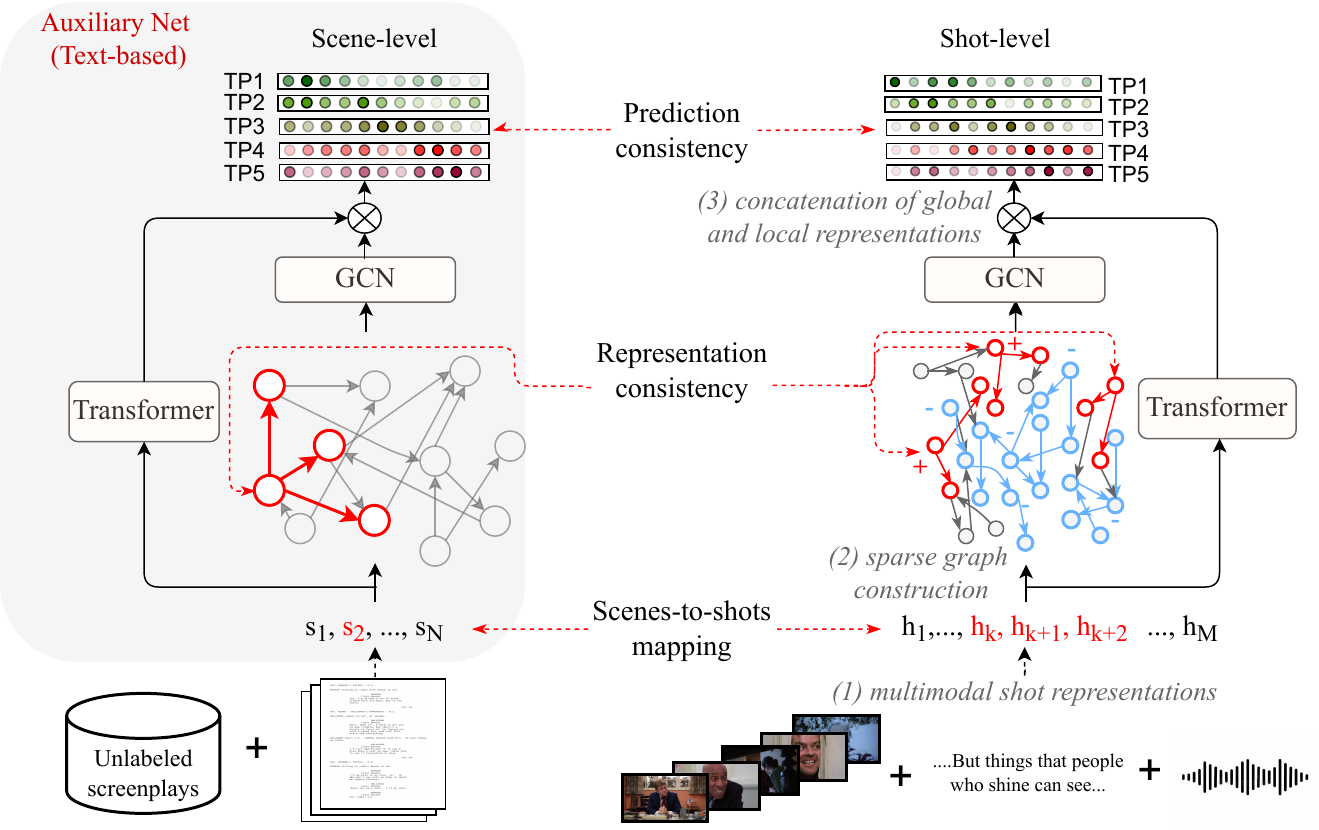}
    \caption{Two networks process \textit{different views} of the 
          movie with \textit{different degrees of granularity}. Our main network (right side) takes as input \textit{multimodal} fine-grained \textit{shot}
          representations based on the movie's video stream. The
          auxiliary text-based network (left side) processes \textit{textual scene}
          representations which are coarse-grained and based on the
          movie's screenplay. The networks are trained jointly on TP
          identification with losses enforcing prediction and
          representation consistency between them.}
    \label{fig:dist_model}
\end{figure*}

In the previous section, we discussed how we can identify important
shot-level trailer moments given movie graph~$\mathcal{G}$ and a set
of shots that act as key events (i.e.,~TPs). We now discuss how we
learn~$\mathcal{G}$ and identify TPs in movies in tandem. We
hypothesize that the training signal provided by TP labels
(i.e.,~narrative structure) also encourages exploring more
fine-grained semantic connections between shot-level events via the
graph that is learned~\cite{papalampidi2020movie}.  A neural network
model first creates~$\mathcal{G}$ that represents relations between
shots in the latent space and then computes the probability $p(y_{it}
|h_i,\mathcal{F}, \theta_1)$, where~$y_{it}$ is a binary label
denoting whether shot~$h_i$ represents TP~$t \in [1,T]$ and $\theta_1$
are network parameters. This network is depicted on the right side of
Figure~\ref{fig:dist_model} and is trained end-to-end on TP
identification.

\vspace{0.5em}
\paragraph{Movie Input} Let~$\mathcal{F}$ denote a full-length movie
consisting of~$M$ shots $\mathcal{F}= \{h_1, h_2, \dots, h_M\}$. For
each shot~$i$, we consider visual (i.e.,~sequence of frames), audio
(i.e.,~audio segments), and textual (i.e.,~subtitles) information and
compute a combination vector $h_i$ of all modalities (see step (1),
right part of Figure~\ref{fig:dist_model}). First, we compute the
textual representation $t_i'$ for the $i^{th}$ shot, via the
bi-directional attention flow \mbox{(\hspace*{-.5ex}\cite{seo2016bidirectional,kim2020dense})}
between subtitles~$t_i$ and audio~$a_i$ and video frames~$v_i$:
\begin{gather}
    S_{t_i, a_i} = a_i^{T}t_i, \quad
    S_{t_i, v_i} = v_i^{T}t_i \\
    a_{i,att} = \operatorname{softmax}(S_{t_i,a_i})a_i, \quad
    v_{i,att} = \operatorname{softmax}(S_{t_i,v_i})v_i \\
    t_i' = a_{i,att} + v_{i,att} + t_i 
\end{gather}
The final
audio~$a_i'$ and visual~$v_i'$ representations are obtained
analogously. Next, vectors $t_i'$, $a_i'$, and $v_i'$ are projected to a lower dimension via a fully-connected linear layer and L2 normalization. Finally, we compute the multimodal representation $h_i$ for shot $i$ via a non-linear projection: \mbox{$h_i = f([t_i'; v_i'; a_i'])$}, where $f(\cdot)$ is a fully-connected layer followed by the ReLU non-linearity.

\vspace{0.5em}

\paragraph{Graph Structure} We  construct a fully-connected
graph described by matrix~$E$. Each cell in this matrix denotes the
similarity~$e_{ij}$ between multimodal shot vectors~$h_i$ and~$h_j$:
\begin{gather}
     e_{ij} = \tanh(W_{i}h_{i} + b_{i})
            \tanh(W_{j}h_{j}+ b_{j}) + b_{ij}
\end{gather}
We then normalize similarities $e'_{ij}$ using the softmax function
(row-wise normalization in matrix $E$). We thus obtain a complete
directed graph, where edge~$e'_{ij}$ records the probability
that~$h_i$ is connected to~$h_j$.  In order to avoid dense connections
which lead to worse contextualization and computational overhead, we
sparsify the graph by selecting a small but variable-length
neighborhood per shot~\cite{papalampidi2020movie}. During
sparsification, we also constrain the adjacency matrix of the graph to
be upper triangular (i.e.,~allowing only future connections between
shots).  We select the top-$k$ neighborhood $\mathcal{P}_i$ per
shot~$h_i$ as~$\mathcal{P}_i =
\operatorname{argmax}\limits_{j\in[1,M],
  |\mathcal{P}_i|=k}e'_{ij}$. Instead of deciding on a fixed number of
$k$~neighbors for all shots and movies, we determine a predefined set
of options for~$k$ (e.g., integers contained in a set $O$) and learn
to select~$k$ via parametrized function:
$z_{i} = \mathrm{softmax}(W_{n}e_{i} + b_{n})$, where~$z_i$ is a
probability distribution over the neighborhood size options for
shot~$h_i$, $W_n \in {\rm I\!R}^{MxO}$, and $e_{i}$ is a vector of
similarities between shot $i$ and all other shots. Hence, the final
neighborhood size for shot~$h_i$ is: $k_i=\operatorname{argmax}_{t\in
  O}z_{it}$ (see step (2) in the right part of
Figure~\ref{fig:dist_model}).  We address discontinuities in our model
(i.e., top-$k$ sampling, neighborhood size selection) by utilizing the
Straight-Through Estimator \cite{bengio2013estimating}. During the
backward pass we compute the gradients with the Gumbel-softmax
reparameterization trick
\mbox{(\hspace*{-.5ex}\cite{maddison2017concrete,jang2017categorical})}.

\vspace{0.5em}

\paragraph{TP Identification} We contextualize
\textit{all shots} with respect to the \textit{entire movie} via a
transformer encoder~\cite{vaswani2017attention} to compute global shot
representations (since transformers can be viewed as fully-connected
graphs); we additionally encode the \textit{graph neighborhood} of
each shot via a one-layer Graph Convolution Network (GCN;
\cite{duvenaud2015convolutional,kearnes2016molecular,kipf2017semi})
for local representations that only depend on a small set of shots
(since the learnt graph is sparse). Finally, we combine global and
local shot representations (see step (3) in the right part of
Figure~\ref{fig:dist_model}) to compute probability~$p(y_{it}
|h_i, \mathcal{F}, \theta_1)$. The network is trained by minimizing
the binary cross-entropy loss per TP (see
Section~\ref{sec:experimental-setup} for details on the dataset we
used). After training, we use graph~$\mathcal{G}$ and predicted TP
shots as input to Algorithm~\ref{algorithm}.

\subsection{Auxiliary Text-based Network} \label{sec:auxiliary_net}

Movie screenplays provide a wealth information in addition to
subtitles, e.g., about characters and their role in a scene, their
actions and emotions. Such information, typically conveyed by lines
describing what the camera sees, is difficult to accurately infer from
video (e.g., via person identification, action recognition, event
localization), while it can be easily mined from the
screenplay. Moreover, corpora of screenplays are relatively easy to
collect leading to orders of magnitude larger datasets compared to
collections of full-length videos.

We exploit screenplays by training our main network (described in the
previous section) together with an \textit{auxiliary text-only}
network. The latter has an architecture similar to the main network,
modulo two key differences: (1)~it only considers \textit{textual}
information and (2)~it processes movies at \textit{scene-level};
although a scene is the smallest unit of a story in a screenplay, it
is far more coarse-grained than shots (a scene may last several
minutes). The auxiliary network thus creates a \textit{scene-level}
graph and estimates \textit{scene-level} probabilities~$q(y_{it}
|s_i,\mathcal{D},\theta_2)$ which quantify the extent to which
scene~$s_i$ corresponds to the $t^\mathit{th}$~TP.
%
We represent scenes with a small transformer encoder
which operates over sequences of sentence vectors. As with our main network, we compute contextualized scene representations; a transformer encoder over the entire screenplay yields global representations, while local ones are obtained with a one-layer GCN over a sparse scene-level graph.


\subsection{Knowledge Distillation}
\label{sec:kd}

We now describe our joint training regime for the two networks which
encapsulate different views of the movie in terms of \textit{data
  streams} (multimodal vs. text-only) and \textit{their segmentation
  into semantic units} (shots vs. scenes), while assuming a
(one-to-many) mapping from screenplay scenes to movie shots (see
Section~\ref{sec:experimental-setup} for details). Traditionally, in
knowledge distillation
\mbox{(\hspace*{-.5ex}\cite{ba2014deep,hinton2015distilling})} the
teacher model (here the auxiliary network) is trained first, and the
knowledge is then asynchronously distilled in a later step to the
student network (here the main network). We propose to
\textit{jointly} train the two networks, since they have complementary
information and can benefit from each other.

\vspace{0.5em}

\paragraph{Prediction Consistency Loss} We aim to enforce some degree
of agreement between the TP predictions of the two networks. For this
reason, we train them jointly and introduce additional constraints in
the loss objective. Similarly to knowledge distillation settings
\mbox{(\hspace*{-.5ex}\cite{ba2014deep,hinton2015distilling})}, we
utilize the KL divergence loss between the auxiliary-based posterior
distribution~$q(y_t|\mathcal{D})$ and the distribution
$p(y_t|\mathcal{F})$ of our main network (upper part in
Figure~\ref{fig:dist_model}).

While in standard knowledge distillation settings both networks
produce probabilities over the same units, in our case, our main network
predicts TPs for shots and the auxiliary one for scenes. We obtain
scene-level probabilities $\overline{p(y_t|\mathcal{F})}$ for the main network by
aggregating shot-level ones via max pooling and re-normalization.  We
then calculate the prediction consistency loss between the two
networks as:
\begin{gather}
\mathcal{P} =
\frac{1}{T}\sum_{t=1}^{T}\mathcal{D}_{KL}\left(\overline{p(y_t|\mathcal{F})}
  \,\,\middle\|\, q(y_t|\mathcal{D}) \right) \label{eq:prediction_consistency}
\end{gather}

\vspace{0.5em}
 
\paragraph{Representation Consistency Loss} We further use a second
regularization loss between the two networks in order to enforce
consistency between the two graph-based representations (i.e.,~over
video shots and screenplay scenes). The purpose of this loss is
twofold: to improve TP predictions for the two networks, as shown in
previous work on contrastive representation learning
\mbox{(\hspace*{-.5ex}\cite{oord2018representation,sun2020contrastive,pan2020spatio})},
and also to help learn more accurate connections between shots (recall
that the shot-based graph serves as input to our graph traversal
algorithm; Section~\ref{sec:graph_traversal}).  In comparison with
screenplay scenes, which describe self-contained events, video shots
are only a few seconds long and rely on surrounding context for their
meaning. We hypothesize that by enforcing the graph neighborhood for a
shot to preserve semantics similar to the corresponding screenplay
scene, we will encourage the selection of appropriate neighbors in the
shot-level graph (middle part of Figure~\ref{fig:dist_model}).

We again first address the problem of varying granularity in the
representations of the two networks. We compute an aggregated
scene-level representation $\overline{h_j}$ based on shots
$h_i, \dots, h_{i+k}$ via mean pooling and calculate the noise
contrastive estimation (NCE;
\cite{gutmann2010noise,wu2018unsupervised}) loss for the
$j^\mathit{th}$ scene:
\begin{gather}
    \mathcal{R} = - \frac{1}{N} \sum_{j=1}^{N} \log
    \frac{e^{s(\overline{h_j}, s_{j})/\tau}}{e^{s(\overline{h_j},
        s_{j})/\tau} + \sum_{\substack{k=1 \\ k\neq j}}^{N}e^{s(\overline{h_j},
        s_k)/\tau}}  \label{eq:standard_contrastive}
\end{gather}
where $N$~is the number of scenes, $s_j$~is the scene representation
calculated by the GCN in the auxiliary network, $\overline{h_j}$~is
the (average) scene representation calculated by the GCN in the main
network, $s(\cdot)$ is a similarity function (here the scaled dot
product), and $\tau$~is a temperature hyperparameter.

\vspace{0.5em}

\paragraph{Joint Training} Our final joint training objective takes
into account the individual losses~$\mathcal{S}$ and~$\mathcal{V}$ of
the auxiliary and main networks, respectively (see Appendix, Section
1.1 for details), and the two consistency losses $\mathcal{P}$ (for
prediction) and $\mathcal{R}$ (for representation):
\begin{gather}
    \mathcal{L}_{TP} = \mathcal{S} + \mathcal{V} + a\mathcal{P} + b\mathcal{R} \label{eq:loss}
\end{gather}
where $a,b$ are hyperparameters modulating the importance of
prediction vs. representation consistency.
Figure~\ref{fig:dist_model} provides a high-level illustration of our
training regime.

\subsection{Self-supervised Pretraining}
\label{sec:pretraining}

We further pretrain the auxiliary network on more \textit{textual} data, which is easier to acquire than videos (e.g.,~fewer
copyright issues and less computational overhead), in order to learn better scene representations. We hypothesize that this knowledge can then be transferred to our main network via the consistency losses in the joint training regime (Equations~\ref{eq:prediction_consistency} and~\ref{eq:standard_contrastive}).


Pretraining takes place on Scriptbase
\cite{gorinski-lapata-2015-movie}, a dataset which consists of
$\sim$1,100 full-length screenplays (approx. 140K~scenes). We adapt a
self-supervised task, namely Contrastive Predictive Coding (CPC;
\cite{oord2018representation}), to our setting: given a
(contextualized) scene representation, we learn to predict a future
representation in the screenplay. We consider a context window of several
future scenes, rather than just one. This is an attempt to account for
non-linearities in the screenplay, which can occur because of unrelated
intervening events and subplots.  Given the representation of an
anchor scene $g_i$, a positive future representation $c_i^{+}$ and a
set of negative examples $\{c_{i1}^{-}, \dots, c_{i(N-1)}^{-}\}$, we
compute the InfoNCE \cite{oord2018representation} loss as:
\begin{gather}
    \hspace{-1em}\mathcal{L}_\mathit{self} = - \frac{1}{N} \sum_{j=1}^{N} \log \frac{e^{s(g_i, c_i^{+})/\tau}}{e^{s(g_i, c_i^{+})/\tau} + \sum_{k=1}^{N-1}e^{s(g_i, c_{ik}^{-})/\tau}} \label{eq:info_nce}
\end{gather}
We obtain scene representations~$g_i$ based on $s_i$ provided by the
one-layer GCN. Starting from the current scene, we perform a random
walk of~$k$ steps and compute~$g_i$ from the retrieved path~$p_i$ in
the graph via mean pooling (more details can be found in the Appendix,
Section 1.2).

\subsection{Sentiment Prediction}
\label{sec:sentiment_prediction}

Finally, our model takes into account how sentiment flows from one
shot to the next. We predict sentiment scores per shot with the same
joint architecture and training regime we use for TP identification
(Sections~\ref{sec:tp_identification}--\ref{sec:kd}). The only
difference for sentiment prediction is in the downstream objective
(losses~$\mathcal{S}$ and~$\mathcal{V}$ in Equation~\ref{eq:loss}).
The main network is trained on shots with sentiment labels
(i.e.,~positive, negative, neutral; cross-entropy loss), while the
auxiliary network is trained on scenes with sentiment labels
(Section~\ref{sec:experimental-setup} explains how the labels are
obtained). After training, we predict a probability distribution over
sentiment labels per shot to capture sentiment flow and discriminate
between high- and low-intensity shots (see Appendix for details).

\section{Experimental Setup}
\label{sec:experimental-setup}

\textbf{Datasets} Our model was trained on TRIPOD$\bigoplus$, an
expanded version of the TRIPOD
dataset~\mbox{(\hspace*{-.5ex}\cite{papalampidi2019movie,papalampidi2020movie})}
which contains 122 screenplays with silver-standard TP annotations
(scene-level)\footnote{{\scriptsize
    https://github.com/ppapalampidi/TRIPOD}} and the corresponding
videos.\footnote{{\scriptsize
    https://datashare.ed.ac.uk/handle/10283/3819}} For each movie, we
further collected as many trailers as possible from YouTube, including
official and (serious) fan-based ones, or modern trailers for older
movies.  To evaluate the trailers produced by our algorithm, we also
collected a new held-out set of~41 movies. These movies were selected
from the
Moviescope\footnote{http://www.cs.virginia.edu/$\sim$pc9za/research/moviescope.html}
dataset \cite{cascante2019moviescope} which contains official movie
trailers. The held-out set does not contain any additional
information, such as screenplays or TP annotations. Detailed statistics of
TRIPOD$\bigoplus$ are presented in Table~\ref{tab:dataset}.

\begin{table}[t]
\centering
\small
\begin{tabular}{@{}l@{~~}r@{}r@{~~}r@{}}
\hline
TRIPOD$\bigoplus$ & \multicolumn{1}{c}{Train} & \multicolumn{1}{c}{Dev+Test} & \multicolumn{1}{c}{Held-out}  \\ \hline
No. movies & 84 & 38 & 41 \\
No. scenes & 11,320 & 5,830 & --- \\
No. video shots & 81,400 & 34,100 & 48,600 \\ 
No. trailers & 277 & 155 & 41 \\
Avg. movie duration & 6.9k (0.6) & 6.9k (1.3) & 7.8k (2.3) \\
Avg. scenes per movie  & 133 (61) & 153 (54) & --- \\
Avg. shots per movie  & 968 (441) & 898 (339) & 1,186 (509) \\
\hfill duration & 6.9 (15.1) & 7.1 (13.6) & 6.6 (16.6) \\
Avg. valid shots per movie & 400 (131) & 375 (109) & 447 (111) \\
\hfill duration & 13.5 (19.0) & 13.9 (18.1) & 13.3 (19.6) \\
Avg. trailers per movie& 3.3 (1.0) & 4.1 (1.0) & 1.0 (0.0) \\
\hfill duration & 137 (42) & 168 (462) & 148 (20) \\
Avg. shots per trailer & 44 (27) & 43 (27) & 57 (28) \\
\hfill duration & 3.1 (5.8) & 3.9 (14.0) & 2.6 (4.0) \\
\hline
\end{tabular}
\caption{Statistics of TRIPOD$\bigoplus$ dataset (for movies,
  screenplays, and trailers);  standard deviation  within
  parentheses, duration in seconds.}
\label{tab:dataset}
\end{table}

\paragraph{Movie Processing} The modeling approach put
forward in previous sections assumes a (one-to-many) mapping from
screenplay scenes to movie shots. We obtain this mapping by
automatically aligning dialogue in screenplays to subtitles that
contain timestamps using Dynamic Time Warping (DTW;
\cite{myers1981comparative,papalampidi2020movie}).  We first segment
the video into scenes based on this mapping, and then segment each
scene into shots using
PySceneDetect.\footnote{https://github.com/Breakthrough/PySceneDetect}
Shots with less than 100 frames in total are too short for processing
and displaying as part of the trailer and are therefore discarded.

Moreover, for each shot we extract visual and audio features. We
consider three different types of visual features: (1)~we sample one
key frame per shot and extract features using ResNeXt-101
\cite{xie2017aggregated} pre-trained for object recognition on
ImageNet \cite{deng2009imagenet}; (2)~we sample frames with a
frequency of~1 out of every 10 frames (we increase this time interval
for shots with larger duration since we face memory issues) and
extract motion features using the two-stream I3D network pre-trained
on Kinetics \cite{carreira2017quo}; and (3)~we use Faster-RCNN
\cite{girshick2015fast} implemented in Detectron2
\cite{wu2019detectron2} to detect person instances in every key frame
and keep the top four bounding boxes per shot which have the highest
confidence alongside with the respective regional representations.  We
project all individual representations to the same lower dimension and
perform L2-normalization. A shot is represented as their sum.  For the
audio modality, we use YAMNet pre-trained on the AudioSet-YouTube
corpus \cite{gemmeke2017audio} to classify audio segments into
521~audio classes (e.g.,~tools, music, explosion); for each audio
segment contained in a scene, we extract features from the penultimate
layer. Finally, we extract textual features
\cite{papalampidi2020movie} from subtitles and screenplay scenes using
the Universal Sentence Encoder (USE; \cite{cer2018universal}).

\paragraph{Trailer Labels} For evaluation purposes, we need to know
which shots in the movie are trailer-worthy. We obtain these binary
labels as follows. We segment the corresponding trailer into shots and
compute for each shot its visual similarity to all shots in the
movie. Movie shots with highest similarity values receive positive
labels (i.e.,~they should be in the trailer).  However, since trailers
also contain shots that are not in the movie (e.g.,~black screens with
text, or material that did not make it into the final cut), we do not
map trailer shots to movie shots unless their similarity is above a
certain threshold (set to 0.85 in our experiments).


\paragraph{Sentiment Labels}
Since TRIPOD does not contain sentiment annotations, we instead obtain
silver-standard labels via COSMIC \cite{ghosal2020cosmic}, a
commonsense-guided framework with state-of-the-art performance for
sentiment and emotion classification in natural language
conversations. Specifically, we train COSMIC on MELD
\cite{poria2019meld}, which contains dialogues from episodes of the TV
series \textit{Friends} and is more suited to our domain than other
sentiment classification datasets
(e.g.,~\cite{busso2008iemocap,li2017dailydialog}).
After training, we use COSMIC to produce sentence-level sentiment
predictions for the TRIPOD screenplays. The sentiment of a scene
corresponds to the majority sentiment of its sentences.  We project
scene-based sentiment labels onto shots using the same one-to-many
mapping employed for TPs.

\section{Results and Analysis}
\label{sec:results}

\subsection{Knowledge Distillation for TP Identification}

Before evaluating the performance of our model on trailer moment
identification, we first investigate whether our joint
training scheme, which distills information from screenplays to
movie videos, improves \textit{TP identification}, which is the task
that our main network is directly trained on. We split the set of
movies with gold-standard scene-level TP labels into development and
test set and select the top~5 (@5) and top~10 (@10) shots per TP in a
movie. As evaluation metric, we use Partial Agreement (PA;
\cite{papalampidi2019movie}), which measures the percentage of TPs for
which a model correctly identifies at least one gold-standard shot
from the 5 or 10 shots selected from the movie (see Appendix for
details).

\begin{table}[t]
\centering
\small
\begin{tabular}{l@{}c@{~~~}c}
\hline 
 & PA@5$\uparrow$ & PA@10$\uparrow$ \\ \hline
 Random (evenly distributed) & 21.67 & 33.44 \\
 Theory position & 10.00 & 12.22 \\
 Distribution position & 12.22 & 15.56 \\
\graphtp~\cite{papalampidi2020movie} & 10.00 & 12.22 \\ \hdashline
\begin{tabular}[c]{@{}l@{}} Ours w/o graph structure \end{tabular} & 22.22 & 33.33 \\ 
\begin{tabular}[c]{@{}l@{}} Ours with graph structure \end{tabular} & 27.78 & 35.56 \\
\hspace{0.1em}+ Auxiliary net, Asynchronous ($\mathcal{P}$)  & 28.89 & 41.11 \\
\hspace{0.1em}+ Auxiliary net, Asynchronous ($\mathcal{P}+\mathcal{R}$)  & 21.11 & 35.56 \\
\hspace{0.1em}+ Auxiliary net, Contrastive Joint ($\mathcal{P}+\mathcal{R}$) & \underline{33.33} & \underline{47.78} \\
\begin{tabular}[c]{@{}l@{}} \hspace{1em} + pre-training \end{tabular} & \underline{\textbf{34.44}} & \underline{\textbf{50.00}} \\
\hline
\end{tabular}
\caption{Model performance on \textit{TP identification} (test
  set). Our method shown with different training
  regimes. Evaluation metric: Partial Agreement (PA) against top~5
  (@5) and top~10 (@10) shots per TP and movie.}
\vspace{-1em}
\label{tab:TP_results}
\end{table}

Table~\ref{tab:TP_results} summarizes our results on the test set.  We
consider the following comparison systems: \textbf{Random} selects
shots from evenly distributed sections (average of 10 runs);
\textbf{Theory} assigns TP to shots according to screenwriting theory
(e.g.,~``Opportunity'' occurs at 10\% of the movie, ``Change of
plans'' at 25\%, etc.); \textbf{Distribution} selects shots based on
their average position in the training data; \textbf{\graphtp} is the
original model of Papalampidi et al.~\cite{papalampidi2020movie}
trained on screenplays (we identify scenes as TPs and then project
scene-level predictions to shots given our scene-to-shots mapping);
\textbf{Ours w/o graph structure} is a base transformer model without
the graph-related information.  We also use several variants of our
own model (\textbf{Ours with graph structure}): without and with the
auxiliary text-based network, trained asynchronously (i.e.,~first
training the auxiliary network and then transferring the knowledge to
the main network) only with the prediction consistency loss
($\mathcal{P}$), both prediction and representation losses
($\mathcal{P}+\mathcal{R}$), and our contrastive joint training
regime.

We observe that our approach outperforms all baselines, and the
equivalent transformer-based model without the graph
information. Although transformers can encode long-range dependencies
between shots, directly encoding sparse connections learned in the
graph is additionally beneficial. Moreover, asynchronous knowledge
distillation via the prediction consistency loss ($\mathcal{P}$)
further improves performance,
whereas adding the representation consistency loss
($\mathcal{P}+\mathcal{R}$) decreases PA.
The proposed training approach (contrastive joint) performs best. We
observe that encouraging the main network to compute representations
close to the (fixed) auxiliary ones hurts performance leading to
information loss, since the two networks capture complimentary
information with visual and audio cues being richer in certain
cases. In contrast, joint training improves both networks'
representations. Finally, pretraining brings further gains, albeit
small, and underlines the benefits of the auxiliary network.

\begin{table}[t]
\centering
\small
\begin{tabular}{L{18em}@{}rr}
\hline
 &  \multicolumn{1}{c}{Dev$\uparrow$} &  \multicolumn{1}{c}{Test$\uparrow$} \\ \hline
Random selection  w/o TPs &  14.47 & 5.61 \\
Random selection with TPs & 20.00 & 9.27 \\ 
\textrank~\cite{mihalcea2004textrank} & 10.26 & 3.66 \\
\graphtrailer w/o TPs & 23.58 & 11.53 \\ 
\graphtrailer with TPs & \textbf{26.95} & \textbf{16.44} \\ \hline
CCANet~\cite{wang2020learning} & 31.05 & 15.12 \\
Supervised \graphtrailer w/o graph & 32.63 & 17.32 \\
Supervised \graphtrailer & \textbf{33.42} & \textbf{17.80} \\
\hline
Upper bound & 86.41 & \multicolumn{1}{c}{---} \\ \hline
\end{tabular}
\caption{Trailer moment identification: performance of unsupervised
  (top) and weakly supervised (bottom) models measured by
  accuracy (\%). All systems create trailers with the same shot budget.}
\vspace{-1em}
\label{tab:trailer_results}
\end{table}

\subsection{Automatic Trailer Moment Identification}

We now evaluate our target task of trailer moment identification on
the held-out set of 41 movies (see Table~\ref{tab:dataset}).
Following prior work~\cite{wang2020learning}, we use accuracy as our
primary evaluation metric (i.e.,~the percentage of correctly
identified trailer shots).  We consider a total budget of 10~shots per
trailer in order to achieve the desired length ($\sim$2~minutes).

We compare \graphtrailer against several unsupervised approaches
(first block in Table~\ref{tab:trailer_results}) including:
\textbf{Random selection} among all shots and among TPs predicted by
our main network, and two \textbf{graph-based systems} based on a
fully-connected graph, where nodes are shots and edges denote the
degree of similarity between them. This graph has no knowledge of TPs,
it is constructed by calculating the similarity between generic
multimodal
representations. \textbf{\textrank}~\cite{mihalcea2004textrank}
operates over this graph to select shots based on their centrality,
while \textbf{\graphtrailer without TPs} traverses the graph with TP
and sentiment criteria removed (Equation~\ref{eq:score}). For the
unsupervised systems which introduce stochasticity and produce
proposals (i.e.,~Random, \graphtrailer), we consider the best trailer
out of 10 proposals. The second block of
Table~\ref{tab:trailer_results} presents supervised approaches which
use trailer labels for training. These include \textbf{CCANet}
\cite{wang2020learning}, which is the state-of-the-art in trailer
moment identification; it considers only visual information and computes
the cross-attention between movie and trailer shots; 
\textbf{Supervised \graphtrailer without graph} is trained for the binary
task of identifying whether a shot should be in the trailer without
considering graph information, screenplays, sentiment or TPs; 
\textbf{Supervised \graphtrailer}  is our main network
(Section~\ref{sec:tp_identification}) trained on trailer moment
identification with trailer-specific labels.

\graphtrailer performs best among unsupervised methods. Interestingly,
\textrank is worse than random, which shows that trailer moment
identification is fundamentally different from vanilla summarization
problems. \graphtrailer without TPs still performs better than
\textrank and random TP selection.\footnote{Performance on the test
  set is lower because we only consider labels from official trailers,
  while the development set contains multiple trailers. We refrained
  from using more trailers for the test set in order to match the
  Moviescope dataset and avoid noise from fan-based videos.}  With
regard to supervised approaches, we find that using all modalities
with a standard architecture (Supervised \graphtrailer w/o graph)
leads to better performance than sophisticated models using visual
similarity (CCANet). By adding graph-related information (Supervised
\graphtrailer), we obtain further improvements.

\begin{table}[t]
\centering
\small
\begin{tabular}{lc}
\hline
 &  {Accuracy} $\uparrow$ \\ \hline
\graphtrailer & 22.63 \\ 
\hspace{0.1em}+ Auxiliary net, Asynchronous ($\mathcal{P}$) & 21.87 \\
\hspace{0.1em}+ Auxiliary net, Asynchronous ($\mathcal{P}+\mathcal{R}$) & 22.11 \\ 
\hspace{0.1em}+ Auxiliary net, Contrast Joint ($\mathcal{P}+\mathcal{R}$) & 25.44 \\ 
 \hspace{1em} + pre-training & \textbf{25.79} \\
\hline
\end{tabular}
\caption{Different training regimes for \graphtrailer: accuracy (\%) of correctly identified trailer
  shots. For direct comparison, \graphtrailer here only uses the
  narrative structure criterion, no information about sentiment
  intensity is considered.} 
\vspace{-1em}
\label{tab:ablation_training}
\end{table}

\begin{table}[t]
\centering
\small
\begin{tabular}{L{16em}c}
\hline
 &  {Accuracy} $\uparrow$ \\
 \hline
Similarity & 24.28 \\
Similarity + TPs & 25.79 \\
Similarity + sentiment & 23.97 \\
Similarity + TPs + sentiment & \textbf{26.95} \\
\hline
\end{tabular}
\caption{\graphtrailer with different
  criteria for performing random walks in the movie graph (Algorithm~1,
  Equation~\eqref{eq:score}).}
\vspace{-1em}
\label{tab:ablation_criteria}
\end{table}

We performed two ablation studies on the development set for
\graphtrailer. We first assessed how the different training regimes of
the dual network influence downstream performance (on the trailer
moment identification task).  Table~\ref{tab:ablation_training} shows
that asynchronous training does not offer any discernible improvement
over the base model. However, when we jointly train the main and
auxiliary networks (using prediction and representation consistency
losses), performance increases by nearly~3\%. A further small increase
is observed when the auxiliary network is pre-trained on more data.

Our second ablation study concerned the criteria used for performing
random walks on the graph~$\mathcal{G}$ (Equation~\ref{eq:score}). As
shown in Table~\ref{tab:ablation_criteria}, when we enforce the nodes
in the selected path to be close to key events (similarity + TPs)
performance improves. When we rely solely on sentiment (similarity +
sentiment), performance drops slightly. This suggests that in contrast
to previous approaches which mostly focus on superficial visual
attractiveness
\mbox{(\hspace*{-.5ex}\cite{xu2015trailer,wang2020learning})} or
audiovisual sentiment analysis \cite{smith2017harnessing}, sentiment
information on its own is not sufficient and may promote outliers that
do not fit well in a trailer. Inspection of trailers produced using
the sentiment criterion alone revealed that it tends to promote shots
that contain intense visuals and audio (e.g.,~music, explosions, etc.)
but are loosely connected to the plot.  When sentiment information is
combined with knowledge about narrative structure (similarity + TPs +
sentiment), trailer shots are not only interesting but also related to
important events in the movie. This further validates our hypothesis
that the two theories about creating trailers are complementary and
can be combined.

Finally, since we have multiple trailers per movie (in the dev set),
we can measure the overlap between their shots (\textbf{Upper
  bound}). The average overlap is 86.14\%, demonstrating good
agreement between trailer makers and a big gap between and automatic
models and human performance.

\begin{table}[t]
\centering
\small
\begin{tabular}{@{}l@{~~~}c@{~~~}c@{~~~}c@{~~~}c@{~~~}r@{}}
\hline
 & \multicolumn{1}{@{}c}{Q1} & \multicolumn{1}{@{}c}{Q2} & Best & Worst & BWS \\
 \hline
Random selection w/o TPs & 38.2 & 45.6 & 19.1 & 25.9 & $-$1.26 \\
\graphtrailer w/o TPs & 37.2 & 44.5 & \textbf{24.4} & 25.9 & $-$0.84 \\
\graphtrailer w/ TPs & \textbf{41.4} & \textbf{48.2} & 20.8 & \textbf{11.6} & \textbf{1.40} \\
\hline
CCANet & 37.7 & 46.6 & 14.3 & 15.2 & $-$0.14  \\
Superv. \graphtrailer & 37.7 & 47.1 & 21.4 & 21.4 & 0.84 \\
\hline
\end{tabular}
\caption{Human evaluation on held-out set. Percentage of Yes
  answers for: Does the trailer contain sufficient information (Q1)
  and is it attractive (Q2). Percentage of times each system was
  selected as Best or Worst, and standardized best-worst scaling
  score.}
  \vspace{-1em}
\label{tab:human_evaluation}
\end{table}

\begin{figure*}[t!]
    \centering
\begin{tabular}{|c|c|c|} \hline
        \begin{tabular}[c]{c} {{\small Step 1 - Initialize path: sample shot}} \\ {{\small from TP1 predictions}} \end{tabular} & \begin{tabular}[c]{c} {{\small Step 2 - Select next shot from}} \\ {{\small immediate neighborhood}} \end{tabular} & \begin{tabular}[c]{c} {{\small Step 3 - Select next shot from}} \\ {{\small immediate neighborhood}} \end{tabular}  \\ 
         \includegraphics[width=0.25\textwidth]{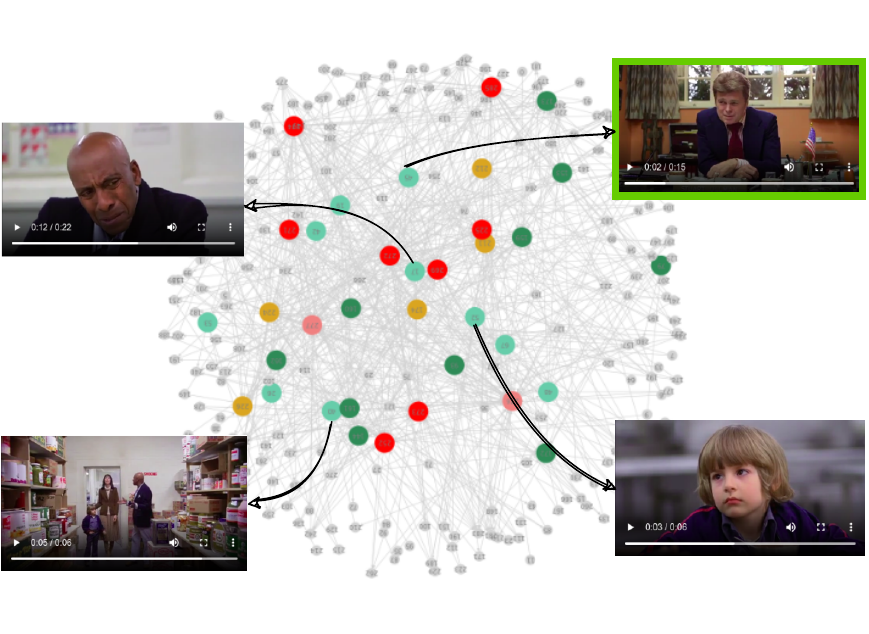} &          
         \includegraphics[width=0.25\textwidth]{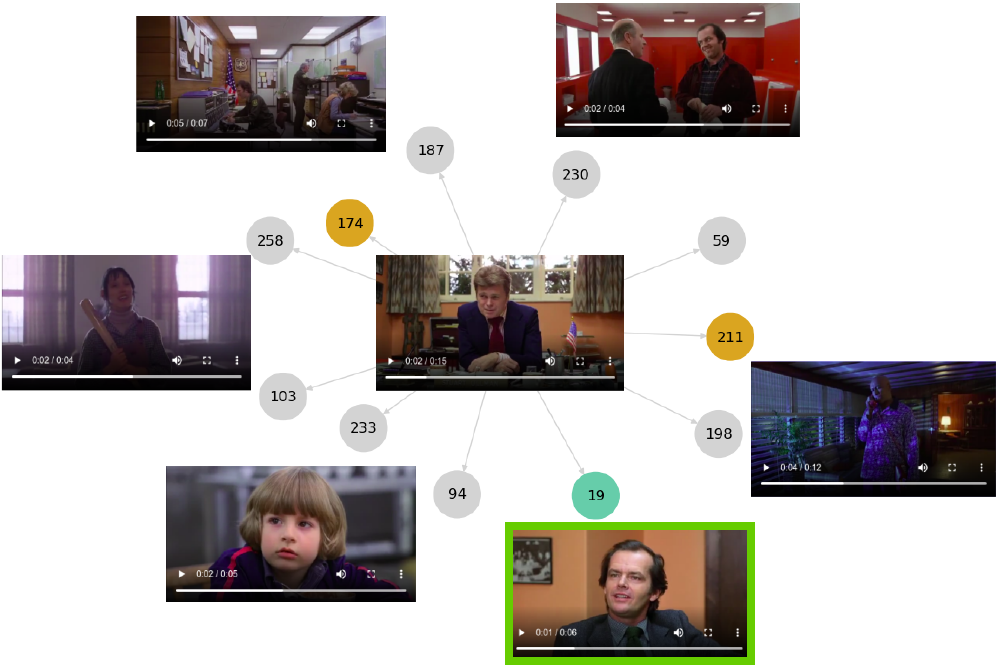} &          
         \includegraphics[width=0.25\textwidth]{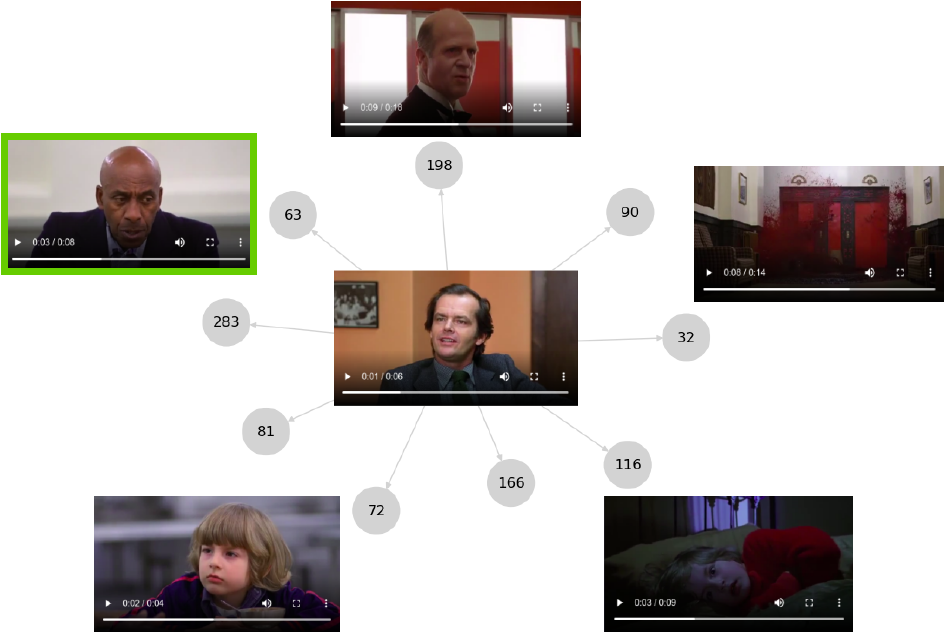} \\ \hline
        
        \begin{tabular}[c]{c} {{\small Step 4 - Select next shot from}} \\ {{\small immediate neighborhood}} \end{tabular} & \begin{tabular}[c]{c} {{\small Step 5 - Select next shot from}} \\ {{\small immediate neighborhood}} \end{tabular} & \begin{tabular}[c]{c} {{\small Final step -  Create trailer based on}} \\ {{\small retrieved sequence of shots}} \end{tabular}  \\ 
         \includegraphics[width=0.25\textwidth]{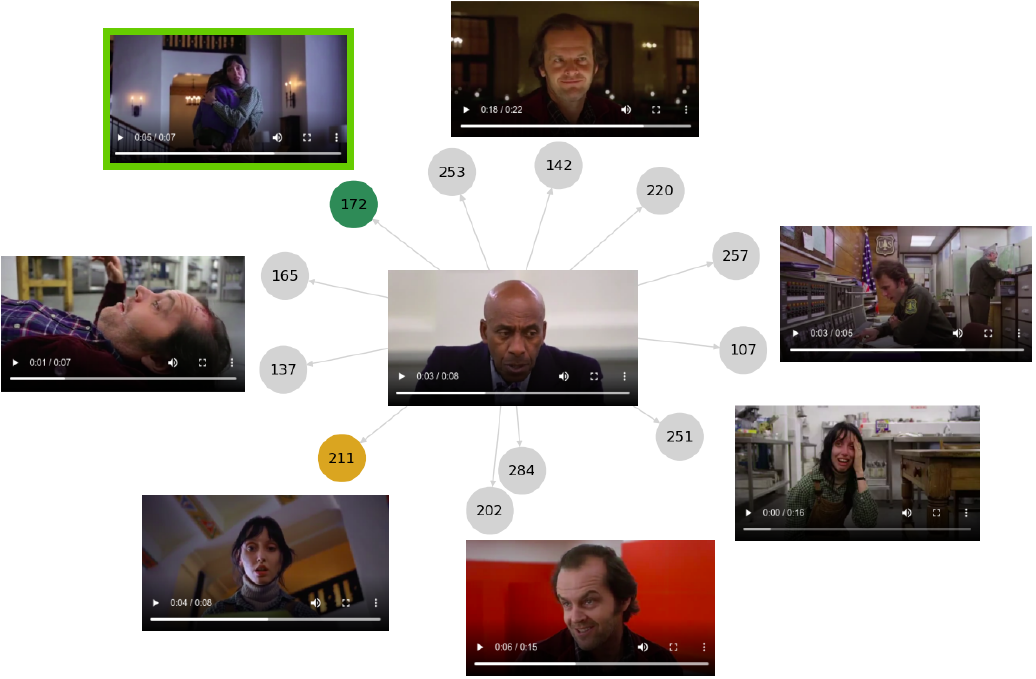} &          
         \includegraphics[width=0.25\textwidth]{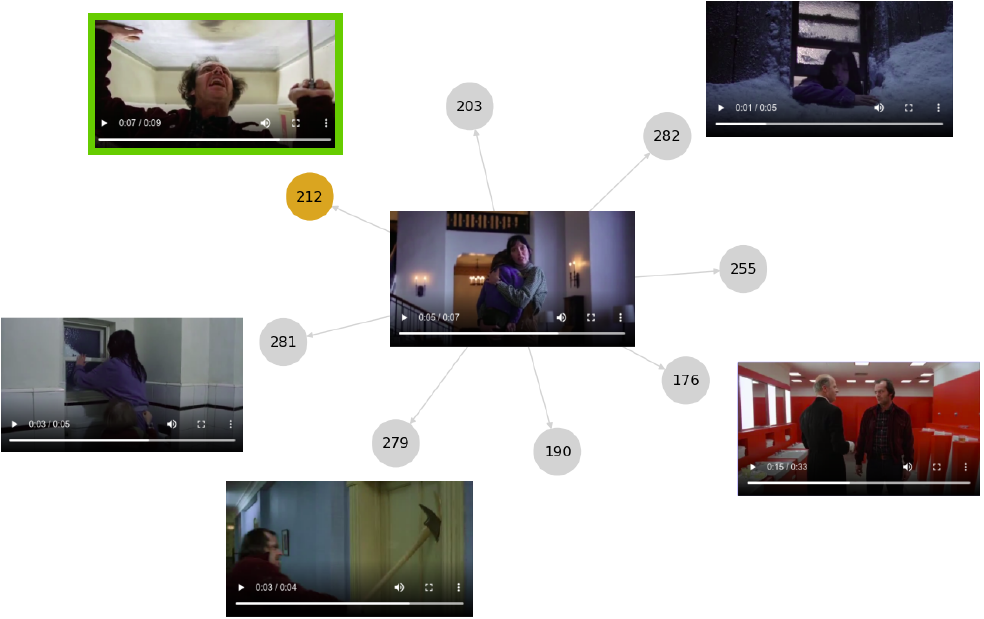} &          
         \includegraphics[width=0.25\textwidth]{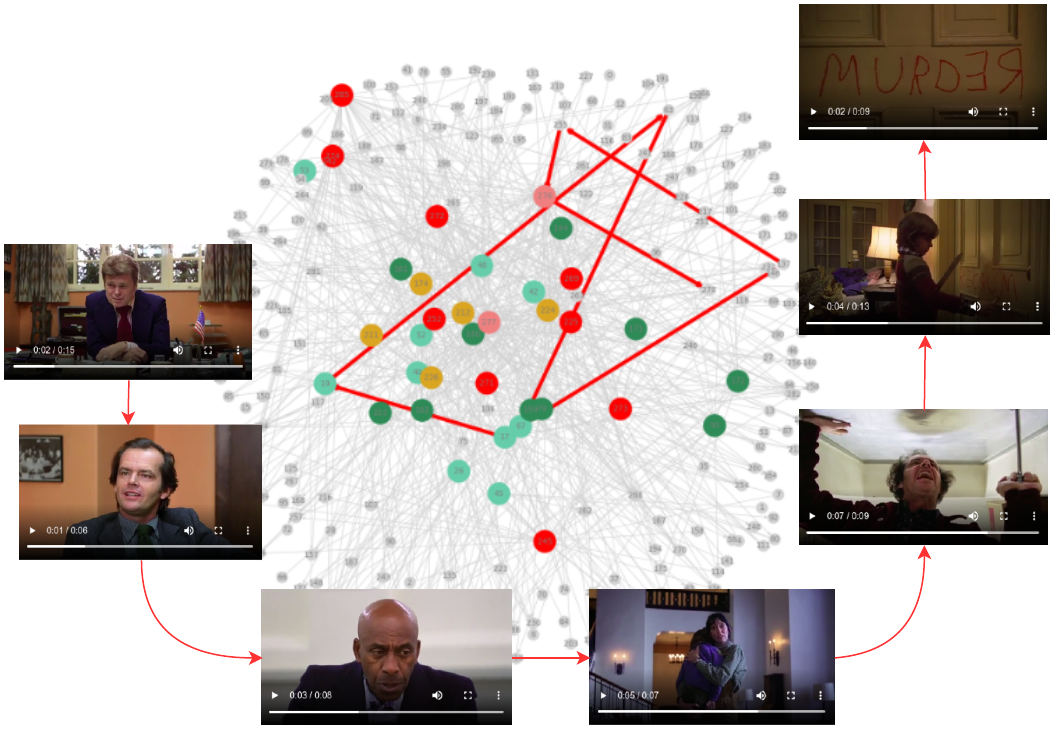} \\ \hline
        
\end{tabular}
 \caption{Run of \graphtrailer algorithm for the movie ``The
   Shining''. Step 1 illustrates the shot-level graph (pruned
   for better visualization) with colored nodes representing the
   different types of TPs predicted in the movie (i.e., {\color{tp1}
     TP1}, {\color{tp2} TP2}, {\color{tp3} TP3}, {\color{tp4} TP4},
   {\color{tp5} TP5}).  Our algorithm starts by sampling a shot
   identified as {\color{tp1} TP1} (Step 1). For each
   next step, we only consider the immediate neighborhood of the
   current shot (i.e.,~6--12 neighbors) and select the next shot based
   on the following criteria: (1) semantic similarity, (2) time
   proximity, (3) narrative structure, and (4) sentiment intensity
   (Steps 2--5 or beyond). Finally, we assemble the proposal trailer (Final step) by
  concatenating the shots in the path. When our algorithm is used as an interactive tool, it allows users to
  review candidate shots at each step and manually select the best
  one while taking into account our criteria.  Users create trailers by only reviewing around 10\%~of the
  movie.
   \label{fig:random_walk_1}}
\end{figure*}

\subsection{Human Evaluation of Trailer Moment Identification} \label{sec:human_eval}

We also conducted a human evaluation study to assess the quality of
the selected trailer shots. For human evaluation, we include Random
selection without TPs as a lower bound, the two best performing
unsupervised models (i.e.,~\graphtrailer with and without TPs), and
two supervised models: CCANet, which is the previous state-of-the-art
for trailer moment identification, and the supervised version of our
model, which is the best performing model according to automatic
metrics.\footnote{We do not include gold-standard trailers in the
human evaluation, since they are post-processed (i.e.,~montage,
voice-over, music) and thus not directly comparable to automatic
ones.}  We generated trailers for all movies in the held-out set by
concatenating the identified trailer shots. We then asked Amazon
Mechanical Turk (AMT) crowd workers to watch all trailers for a movie,
answer questions relating to the information provided (Q1) and the
attractiveness (Q2) of the trailer, and select the best and worst
trailer (see details in Appendix, Section~6). We collected assessments from five
different judges per movie.

Table~\ref{tab:human_evaluation} shows that \graphtrailer with TPs
provides on average more informative (Q1) and attractive (Q2) trailers
than all other systems (pairwise differences are significant using a $\chi^{2}$ test). Although \graphtrailer without TPs and
Supervised \graphtrailer are more often selected as best, they are
also chosen equally often as worst. When we compute standardized
scores (z-scores) using best-worst scaling \cite{bestworst}, \graphtrailer with TPs
achieves the best performance (note that is also rarely selected as
worst) followed by Supervised \graphtrailer. Interestingly,
\graphtrailer without TPs is most often selected as best (24.40\%),
which suggests that the overall approach of modeling movies as graphs
and performing random walks instead of individually selecting shots
helps create coherent trailers. However, the same model is also most
often selected as worst, which shows that this naive approach on its
own cannot guarantee good-quality trailers. 

\paragraph{Spoiler Alert} Our model does not \emph{explicitly}
avoid spoilers when selecting trailer shots.  We experimented with a
spoiler-related criterion when traversing the movie graph in
Algorithm~\ref{algorithm}. Specifically, we added a penalty when
selecting shots that are in ``spoiler-sensitive'' graph
neighborhoods. We identified such neighborhoods by measuring the
shortest path from the last two TPs, which are by definition the
biggest spoilers in a movie. However, this variant of our algorithm
resulted in inferior performance according to automatic metrics and we thus did not pursue it
further. We believe that such a criterion is not beneficial for
selecting trailer shots, since it discourages the model from
selecting exciting shots from the latest parts of the movie. These
high-tension shots are important for creating interesting trailers and
are indeed included in real-life trailers. More than a third of
professionally created trailers in our dataset contain shots from the last two
TPs (``Major setback'', ``Climax''). We discuss this further in
the Appendix (see Sections 4 and 5).


\section{\graphtrailer as an Interactive Tool}
\label{sec:discussion}

One of the advantages of our algorithm is that it uses interpretable
shot selection criteria which can be easily turned on and off depend
on user preferences.  In the following, we describe how our algorithm
works via an example, highlighting its human-in-the-loop functionality
(Section~\ref{sec:interactive_description}). We also provide a
use-case where trailers are created semi-automatically using
\graphtrailer\ proposals (Section~\ref{sec:demo_results}).

\begin{figure}[t]
    \centering
    \includegraphics[width=\columnwidth]{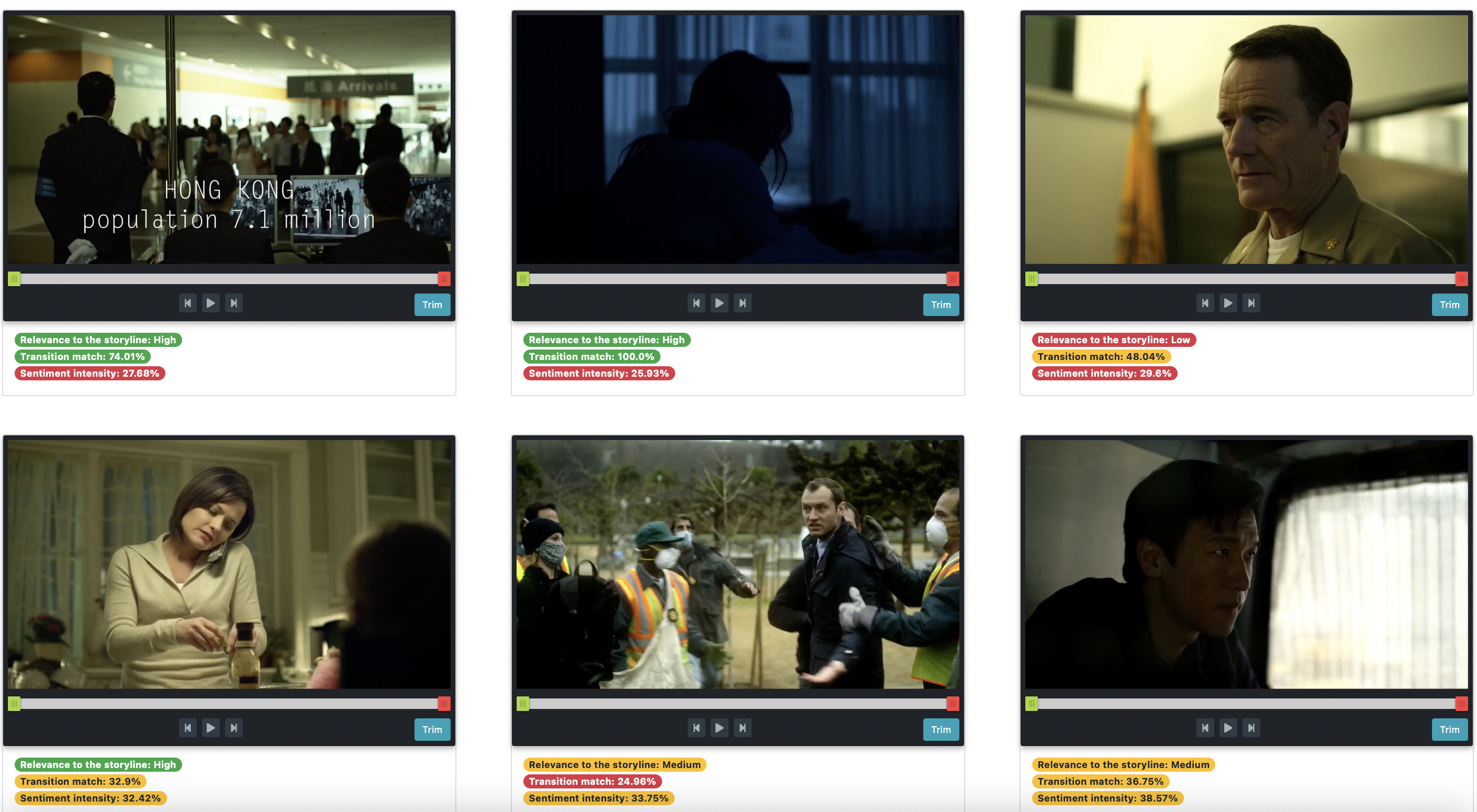}
    \caption{Users can manually review a limited set of shots to be
      included in the trailer given metadata, such as importance,
      sentiment intensity, and transition match; they create trailers
      interactively step by step by selecting and (optionally)
      trimming best shots.}
    \label{fig:demo_example}
\end{figure}

\subsection{Method} \label{sec:interactive_description}

We present in Figure~\ref{fig:random_walk_1} an example of how
\graphtrailer operates over a sparse (shot-level) graph for the movie ``The Shining''.
We begin with shots that have been identified as TP1
(i.e.,~``Opportunity''; introductory event for the story). We sample a
shot (bright green nodes in graph) and initialize our path. For the
next steps (2--5; in reality, we execute up to 10~steps, but we omit a
few for the sake of brevity), we only examine the immediate
neighborhood of the current node and select the next shot to be
included in the path based on the following criteria: (1)~semantic
coherence, (2)~time proximity, (3)~key events, and (4)~sentiment
intensity (see details about how we formalize these criteria in
Section~\ref{sec:graph_traversal}).  We observe that our algorithm
manages to stay close to important events (colored nodes) while
creating the path, which means that we reduce the probability of
selecting random shots with no relevance to the main story. In ``Final
Step'' (Figure~\ref{fig:random_walk_1}), we assemble the trailer
sequence by concatenating all selected shots in the path. We also
illustrate the path in the graph (i.e.,~red line; ``Final Step'' of
Figure~\ref{fig:random_walk_1}).

A similar procedure is followed when \graphtrailer is used as an
interactive tool.  (see our
demo\footnote{https://movie-trailers-beta.herokuapp.com} and the
example in Figure~\ref{fig:demo_example}). Specifically, given the
immediate neighborhood at each step (e.g.,~Step 2 in
Figure~\ref{fig:random_walk_1}), a user selects a shot for the trailer
given a small set of options with corresponding metadata (i.e.,~key
events, sentiment intensity, semantic coherence), which are easy to
review. Moreover, the user can also decide when to finish the trailer,
and move back and forth depending on the set of options that are
presented later in the creation process. Finally, when selecting a
shot, it is possible to trim it to adjust imperfections that might
appear due to automatic shot segmentation. We provide more details and
examples of our interactive tool in the Appendix (Section~7).
Overall, our approach drastically reduces the amount of shots that
need to be reviewed during trailer creation to 10\%~of the movie.
Moreover, our criteria allow users to explore different sections of
the movie, and create diverse trailers.

\begin{figure}[t]
\centering
  \begin{tikzpicture}[scale=.85]
  \begin{axis}[
    xlabel={Relative difference for percentage of  Yes-answers},
    xbar,
    xmin=-50,xmax=70,
    xticklabel style={rotate=60, anchor=east},
     xmajorgrids = true,
    enlarge y limits  = 0.2,
    enlarge x limits  = 0.02,
    axis lines*=left,
    xtick={-40,-30,...,100},
     symbolic y coords = {Preference, Informativeness, Attractiveness,
       Spoilers},
     legend cell align = left,
        legend image code/.code={
        \draw [#1] (0cm,-0.1cm) rectangle (0.2cm,0.25cm); },
      nodes near coords,
     every node near coord/.append style={font=\small},
          legend style={
            at={(-0.2,1.15)}, draw=none, legend columns=-1,
                 anchor=north west,
                  column sep=1ex
          }
  ]
  \addplot[style={Dandelion, fill=Dandelion}] coordinates { (60,Preference)         (15,Informativeness)
                         (33,Attractiveness)  (-20,Spoilers) };
  \addplot[style={RedOrange, fill=RedOrange}] coordinates { (52,Preference)         (-4,Informativeness)
                         (0,Attractiveness)   (40,Spoilers)  };
 \legend{vs.~\graphtrailer, vs.~Gold-standard Selection}
  \end{axis}
\end{tikzpicture}
\vspace*{-.2cm}
  \caption{Human evaluation on semi-automatic trailers created using our interactive tool with a human in the loop. We compare the semi-automatic trailers against fully-automatic ones created by \graphtrailer and a gold-standard selection derived from aligning real trailer shots with movie shots.}
    \label{fig:semi_automatic_human_eval}
\end{figure}

\subsection{Semi-automatic Trailer Creation} \label{sec:demo_results}

We next evaluate whether our interactive tool can provide good quality
outputs while minimizing the time a user spends on the task. By
semi-automatically creating trailers, we can easily correct mistakes of
the automatic process (e.g.,~avoid unwanted spoilers, better combine
shots), while minimizing the time required. We measure this trade-off
between automatic methods and human involvement by comparing 
shots selected via our interactive tool against those selected by
 an expert or an automatic system.

For assembling trailers semi-automatically, we recruited two
non-experts and asked them to first familiarize themselves with the tool
and then select sequences of shots for inclusion in a trailer for all
movies in our held-out set (41~movies in total). Users were encouraged
to select shots by freely going back and forth until they were
satisfied with the result, without however devoting \emph{more than 30
minutes} to each movie. On average, they spent 22~minutes per movie (going
backwards and forward three times). 

To assess the quality of the semi-automatic trailers, we conducted a
human evaluation study similar to the one presented in
Section~\ref{sec:human_eval}. Specifically, Crowdworkers were asked to
watch a pair of trailers for the same movie and indicate which one was
most informative (Q1), attractive (Q2), whether it contained spoilers
(Q3), and which one was best overall.  Participants compared
semi-automatic trailers created as discussed above against fully
automatic ones obtained from \graphtrailer\ or gold-standard labels
(see Section~\ref{sec:experimental-setup}).  We elicited preferences
from five different judges per movie who provided judgments for the
following combinations: (1)~semi-automatic trailers vs. \graphtrailer,
and (2)~semi-automatic trailers vs. gold-standard selection.

We present the results of the human evaluation study in
Figure~\ref{fig:semi_automatic_human_eval} as the relative difference
between semi-automatic trailers and either \graphtrailer or
gold-standard selection.  We observe that AMT participants prefer
semi-automatic trailers against \graphtrailer 60\%~of the time,
indicating that having a human in the loop increases the quality of
trailer shot selection. Moreover, semi-automatic trailers are also
preferred 52\%~of the time against gold-standard selection, suggesting
that the quality of the trailers for both methods is comparable, even
though we instructed our users to make a trailer in less than
30~minutes!

Finally we also present the relative difference between methods for
the percentage of Yes-answers to Q1 (Informativeness), Q2
(Attractiveness), and Q3 (Spoilers). We observe that the
semi-automatic approach increases both the informativeness and
attractiveness of the trailers, while reducing spoilers
by~20\%. Semi-automatic and gold-standard selection are comparable in
terms of informativeness and attractiveness. Semi-automatic trailers,
however, contain 40\%~more spoilers than gold-standard ones. Although
this could be improved in the future, it does not seem to
significantly affect how judges overall rate semi-automatic trailers,
i.e., they still prefer them to \graphtrailer\ and consider them as
good as gold-standard trailers which, incidentally, also contain
spoilers (5.85\%~of the time).

\section{Conclusions}

In this work, we proposed an approach to trailer moment identification
which adopts a graph-based representation of movies and uses
interpretable criteria for selecting shots.  We have also shown that
privileged information from screenplays can be leveraged via
contrastive learning, resulting in a model that identifies turning
points \emph{and} trailer moments (see Appendix, Section~6 for
discussion on how TPs generalize to other tasks and
datasets). Finally, we showcased how our algorithm can be converted
into an assistive tool for trailer creation with a human in the
loop. Semi-automatic trailer creation drastically reduces the number
of shots that need to be reviewed, leading to better quality output
compared to fully automatic methods.

In the future we would like to focus on methods for predicting
fine-grained emotions (e.g.,~grief, loathing, terror, joy) in movies.
In this work, we consider positive/negative sentiment as a stand-in
for emotions, due to the absence of in-domain labeled
datasets. Previous efforts have focused on tweets
\cite{abdul-mageed-ungar-2017-emonet}, YouTube opinion videos
\cite{bagher-zadeh-etal-2018-multimodal}, talkshows
\cite{Grimm:ea:2008}, and recordings of human interactions
\cite{BILAKHIA201552}. Preliminary experiments revealed that
transferring fine-grained emotion labels from other domains to ours leads to
unreliable predictions compared to sentiment which is more stable and
improves trailer moment identification. Avenues for future work
include emotion datasets for movies and emotion detection models based
on textual \emph{and} audiovisual cues.

\section{Ethics Statement}

The human evaluation studies using Mechanical Turk reported in
Sections~\ref{sec:human_eval} and~\ref{sec:demo_results} were approved
by the IRB of the US Air Force Research Laboratory, protocol number
FWR20180142X.  The human-in-the-loop study reported in
Section~\ref{sec:demo_results} was conducted by two employees of the
University of Edinburgh. The School of Informatics ethics committee
confirmed that such studies are exempt from ethical approval.






\ifCLASSOPTIONcaptionsoff
  \newpage
\fi



\bibliographystyle{IEEEtran}
\bibliography{references}
\vspace{-4em}
\begin{IEEEbiographynophoto}{Pinelopi Papalampidi}
is a PhD student in the School of Informatics at the University of Edinburgh and member of EdinburghNLP working on structure-aware narrative understanding and summarization from multiple views. 
\end{IEEEbiographynophoto}
\vspace{-4em}
\begin{IEEEbiographynophoto}{Frank Keller}
is a professor in the School of Informatics at the University of Edinburgh. He is affiliated with EdinburghNLP, the Natural Language Processing Group at the University of Edinburgh. His research focuses on how people solve complex tasks such as understanding language or processing visual information.
\end{IEEEbiographynophoto}
\vspace{-4em}
\begin{IEEEbiographynophoto}{Mirella Lapata}
is professor in the School of Informatics at the University of
Edinburgh.  Her research focuses on computational models for the
representation, extraction, and generation of semantic information
from structured and unstructured data, involving text and other
modalities such as images, video, and large scale knowledge bases.
\end{IEEEbiographynophoto}






\makeatletter
\newcommand*{\radiobutton}{%
  \@ifstar{\@radiobutton0}{\@radiobutton1}%
}
\newcommand*{\@radiobutton}[1]{%
  \begin{tikzpicture}
    \pgfmathsetlengthmacro\radius{height("X")/2}
    \draw[radius=\radius] circle;
    \ifcase#1 \fill[radius=.6*\radius] circle;\fi
  \end{tikzpicture}%
}
\makeatother

\appendices
\clearpage

\section{Model Details}\label{sec:model_details}

In this section we provide details on the various modeling components
of our approach. We begin by discussing how the TP identification
network is trained (Section~\ref{sec:train-tp-ident}), and then give
technical details about pre-training on screenplays
(Section~\ref{sec:pre_training_appendix}), and how sentiment flow is
used for graph traversal (Section~\ref{sec:sent-pred-desir}).

\subsection{Training on TP Identification}
\label{sec:train-tp-ident}

Section 3.4 of the main paper presents our training regime for the
main and auxiliary networks under the assumption that scene-level TP
labels are available (i.e., binary labels indicating whether a scene
acts as a TP in a movie). Given such labels, our model is trained with
a \textit{binary cross-entropy} (BCE) objective between the few-hot
gold labels and the network's TP predictions.

However, in practice, our training set contains \textit{noisy, silver-standard} labels
for scenes. The latter are released together with the TRIPOD
\cite{papalampidi2019movie} dataset and were created automatically.
Specifically, TRIPOD provides \emph{gold-standard} TP annotations for
synopses (not screenplays), under the assumption that synopsis
sentences are representative of TPs. And sentence-level annotations
are projected to scenes with a matching model trained with teacher
forcing \cite{papalampidi2019movie} to create silver-standard labels.

In this work, we consider the probability distribution over screenplay
scenes as produced by the teacher model for each TP and compute the KL
divergence loss between the teacher posterior distributions $q(y_t)$
and the ones computed by our model~$p(y_t)$:
\begin{gather}
 \mathcal{O}_t = \mathcal{D}_{KL}\left(p(y_t) \middle\| q(y_t) \right), t \in [1,T] \label{eq:KL_TPs}
\end{gather}
where $T$ is the number of TPs. Hence, loss~$\mathcal{S}$
(Equation~(9), Section~3.4 of the main paper) used for the auxiliary
network is $\mathcal{S}=\sum_{t=1}^T\mathcal{O}_t$ . Accordingly, for
the main network which operates over shots, we first aggregate shot
probabilities to scene-level ones (see Sections~3.3 and~4 of the main
paper) and then compute the KL divergence between aggregated
probabilities and the teacher distribution (loss $\mathcal{V}$ in
Equation (9), Section 3.4 of the main paper), in the same fashion as
in Equation \eqref{eq:KL_TPs} of the Appendix.

\subsection{Self-supervised Pre-training} 
\label{sec:pre_training_appendix}

In Section~3.5 of the main paper, we describe a modified approach to Contrastive
Predictive Coding (CPC; \cite{oord2018representation}) for
pre-training the auxiliary text-based network on a larger corpus of screenplays
(i.e.,~Scriptbase \cite{gorinski-lapata-2015-movie}). The training
objective is InfoNCE (see Equation~(10), Section~3.5 of the main paper) and
considers structure-aware representations for each scene~$s_i$ in the
screenplay.

Here, we explain in more detail how these representations are
computed. Given scene~$s_i$, we select the next node in the path based
on the edge weights connecting it to its neighbors
$s_t=\operatorname{argmax}\limits_{j\in \mathcal{P}_i}e'_{ij}$ where
$\mathcal{P}_i$ is the sparsified immediate neighborhood for $s_i$. We
perform $k$ such steps and create path $p_i$ which contains
(graph-based) representations (via one-layer GCN) for the selected
nodes. Finally, representation $g_i$ for scene $s_i$ is computed from
this path $p_i$ via mean-pooling.

Notice that we compute structure-aware scene representations~$g_i$ via
random walks rather than stacking multiple GCN layers.  We could in
theory stack two or three GCN layers and thus consider many more
representations (e.g.,~100 or 1,000) for contextualizing~$g_i$.
However, this would result to over-smoothed representations, that
converge to the same vector \cite{oono2019graph}. Moreover, when
trying to contextualize such large neighborhoods in a graph, the
bottleneck phenomenon prevents the effective propagation of long-range
information \cite{alon2020bottleneck} which is our main goal. Finally,
in ``small world'' networks with a few hops, neighborhoods could end
up containing the majority of nodes in the graph
\cite{barcelo2019logical}, which again hinders meaningful exploration
of long-range dependencies. Based on these limitations of GCNs, we
perform random walks, which allow us to consider a small number of
representations when contextualizing a scene, while also exploring
long-range dependencies.

\subsection{Sentiment Flow in \graphtrailer}
\label{sec:sent-pred-desir}

One of the criteria for selecting the next shot in our graph traversal
algorithm (Section~3.1 of the main paper) is the sentiment flow of the
trailer generated so far. Specifically, we adopt the
hypothesis\footnote{https://www.derek-lieu.com/blog/2017/9/10/the-matrix-is-a-trailer-editors-dream}
that trailers are divided into three sections based on sentiment
intensity. The first section has medium intensity for attracting
viewers, the second section has low intensity for delivering key
information about the movie, and finally the third section displays
progressively higher intensity for creating cliffhangers and
excitement for the movie.
 
Accordingly, given a budget of $L$~trailer shots, we expect the
first~$L/3$ ones to have medium intensity without large variations
within the section (e.g., we want shots with average absolute
intensity close to 0.7, where all scores are normalized to a range
from $-$1 to 1). In the second part of the trailer (i.e.,~the next $L/3$
shots) we expect a sharp drop in intensity and shots within this
section to maintain more or less neutral sentiment (i.e.,~0
intensity). Finally, for the third section (i.e.,~the final $L/3$
shots) we expect intensity to steadily increase. In practice, we
expect the intensity of the first shot to be~0.7 (i.e.,~medium
intensity), increasing by 0.1 with each subsequent shot until
we reach a peak at the final shot.

\section{Implementation Details}

\textbf{Evaluation Metrics} Previous work \cite{papalampidi2019movie}
evaluates the performance of TP identification models in terms of
three metrics defined over scenes: Total Agreement (TA), i.e., the
percentage of TP scenes that are correctly identified, Partial
Agreement (PA), i.e., the percentage of TP events for which at least
one gold-standard scene is identified, and Distance (D), i.e., the
minimum distance in number of scenes between the predicted and
gold-standard set of scenes for a given TP, normalized by the
screenplay length. In this work we report results using the partial
agreement metric applied to shots. We can no longer use total
agreement, the metric was devised for scene-level TP detection, for
which gold TP labels were provided. For our task, we only have access
to silver standard shot labels, and assume all shots within a scene
are equally important.  We do not use the distance metric either since
it yields very similar outcomes and does not discriminate among model
variants.

\vspace{0.5em}

\paragraph{Hyperparameters} Following previous work
\cite{papalampidi2020movie}, we project all types of features
(i.e.,~textual, visual, and audio) to the same lower dimension
of~128. We find that larger dimensions increase the number of
parameters considerably and yield inferior results possibly due to
the small dataset size.

We contextualize scenes (with respect to the screenplay) and shots
(with respect to the video) using transformer encoders. We
experimented with 2, 3, 4, 5, and 6 layers in the encoder and obtained
best results with 3~layers. For the feed forward (FF) dimension, we
experimented with both a standard size of~2,048 and a smaller size
of~1,024 and found the former works better. We use another transformer
encoder to compute the representation of a scene from a sequence of
input \emph{sentence} representations. This encoder has 4 layers and
1,024 FF dimension.  Both encoders, employ~8 attention heads and
0.3~dropout.

During graph sparsification (i.e.,~selection of top-$k$ neighbors), we
consider different neighborhood options for the scene- and shot-based
networks due to their different granularity and size. Following
\cite{papalampidi2020movie}, we consider one to six neighbors for
the scene network and we increase the neighborhood to six to 12 for
the shot network.

Overall, the main network has 1.6M learnable parameters and the
auxiliary one has 2.1M parameters. We pre-train the auxiliary network
on a single P100 GPU with 12GB memory for a maximum of 50 epochs,
which approximately takes 7 hours.  We train all variants of our
method including the joint main-auxiliary network variant on a single
P100 GPU with 12GB memory for 30 epochs, which approximately takes 1
hour.

For training our dual network on TP identification, we use the
$\mathcal{L}_{TP}$ objective described in Equation~(9) (Section~3.4 of
the main paper). We set hyperparameters~$a$ and~$b$ that determine the
importance of the prediction and representation consistency losses in
$\mathcal{L}_{TP}$ to 10 and 0.03, respectively.  Moreover, while
pre-training the auxiliary network on the Scriptbase corpus
\cite{gorinski-lapata-2015-movie} we select as future context window
10\% of the screenplay. As explained in Section~3.5 of the main paper
and Section~\ref{sec:pre_training_appendix} of the Appendix, we
compute structure-aware scene representations by performing random
walks of~$k$ steps on the graph starting from an anchor scene. We
empirically choose 3~steps.

Finally, as described in Section~3.1 of the main paper, \graphtrailer uses a combination
of multiple criteria for selecting the next node to be included in the
trailer path. The criteria are combined using hyperparameters which
are tuned on the development set. The search space for each
hyperparameter~$\lambda$ (Equation~2, Section~3.1 of the main paper) is 0, 1, 5, 10,
  15, 20, 25, 30, and we find that the best combination for
$\lambda_1$ (semantic similarity), $\lambda_2$ (time proximity),
$\lambda_3$ (narrative structure), and $\lambda_4$ (sentiment intensity)
is 1, 5, 10, 10, respectively. 

\paragraph{Shot Segmentation} We use
PySceneDetect\footnote{https://github.com/Breakthrough/PySceneDetect}
to segment the movie into shots as mentioned in Section~4 of the main
paper. This tool has different options and settings depending on the
style of video to segment. After experimentation, we used the
``detect-content'' mode, as it provided more accurate segmentation for
a subset of the movies in the training set. Shot boundary detection is
determined by a threshold parameter value; we generated a statistics
file for our videos (training set) and experimented with different
values while empirically evaluating the resulting segmentation on a
diverse subset of movies from the training set. We then selected the
value that (on average) seemed to perform well across different
movies. However, there are still imperfections in the automatic shot
segmentation since we are using a general-purpose tool for widely
varying movies both in terms of style and content.  The interactive
tool we propose allows users to correct imperfections when selecting
content for the trailer as a form of post-processing. We leave better
shot segmentation methods to future work.

\begin{table}[t]
\centering \small
\begin{tabular}{l@{}r@{~~~}r}
\hline 
 & PA@5$\uparrow$ & PA@10$\uparrow$ \\ \hline
Ours (Contrastive Joint + pre-training) & \textbf{34.44} & \textbf{50.00} \\
\hspace{1em} Subtitles only & 21.11 & 36.67 \\
\hspace{1em} Video only & \underline{33.33} & \underline{46.67} \\
\hspace{1em} Audio only & 27.78 & 44.44 \\
\hline
\end{tabular}
\caption{Ablation study on the contribution of different modalities (i.e.,~text, video, audio) on TP identification. Evaluation metric: Partial Agreement (PA) against top~5
  (@5) and top~10 (@10) selected shots per TP and movie. The best performance is marked with bold and the second best is underlined.}
\label{tab:TP_ablation}
\end{table}

\begin{figure}[t]
    \centering
    \includegraphics[width=0.8\columnwidth]{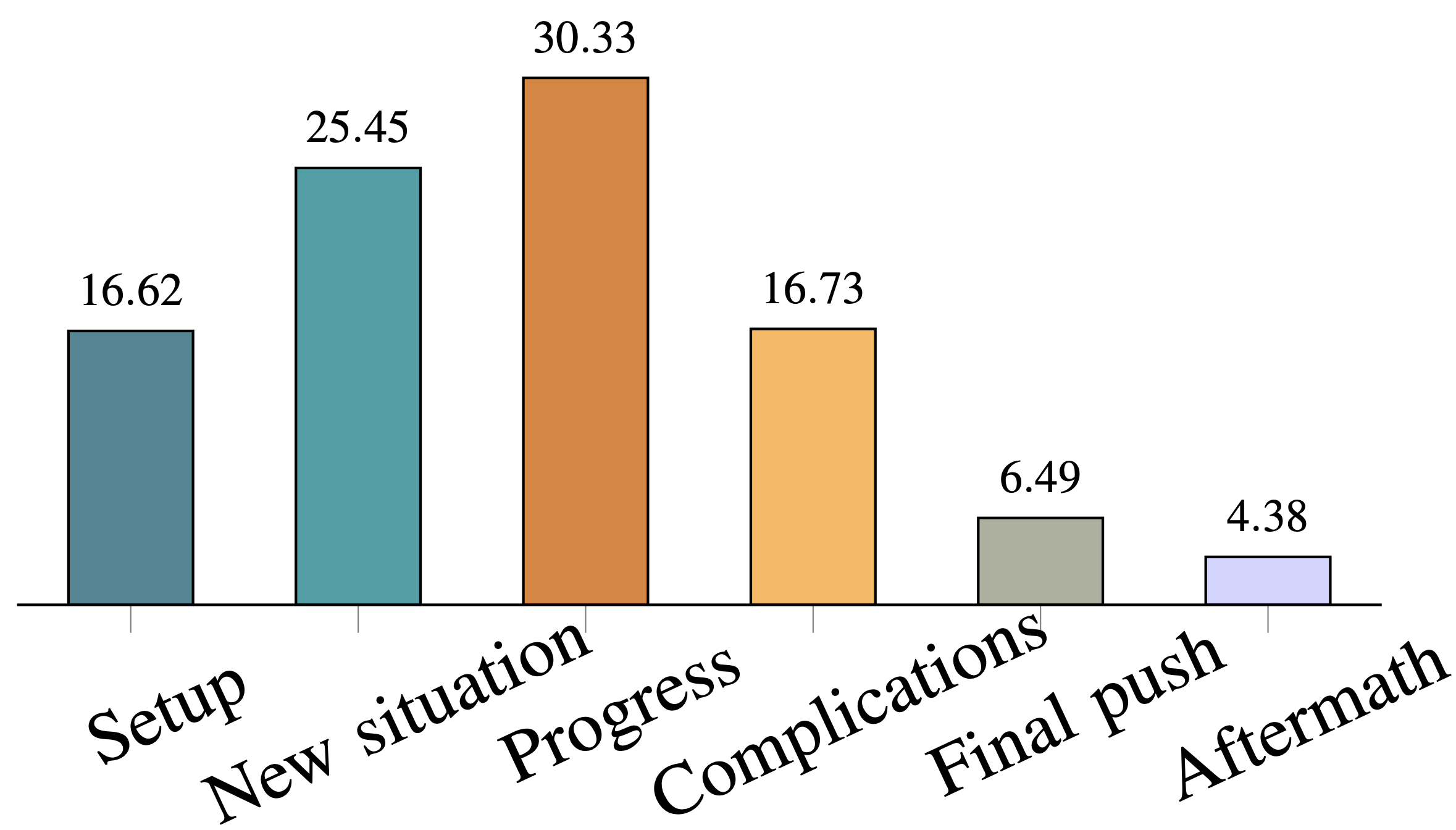}
    \caption{Distribution of trailer shots corresponding to different
      sections of a movie (development set) as determined by TPs.
      Trailer shots come from all parts of the movie, even from the
      end, although the majority are from the beginning and 
      middle.}
    \label{fig:tps_distribution}
\end{figure}

\begin{table}[t]
\centering
\small
\begin{tabular}{L{18em}r}
\hline
Opportunity & 52.63 \\
Change of plans & 55.26 \\
Point of no return & 47.37 \\
Major setback & 34.21 \\
Climax & 34.21 \\
\hline
\end{tabular}
\caption{Proportion of trailers that include at least one shot labeled
  as a specific type of TP on the development set. The first two TPs
  (that present an introduction to the story) appear more frequently
  in trailers, especially in comparison to the last two, which often
  contain major spoilers.}
\label{tab:trailers_vs_trailers}
\end{table}

\begin{table}[t]
\centering
\small
\begin{tabular}{L{12em}c}
\hline
 &  Sentiment Intensity \\
 \hline
First part & 11.50 \\
Second part & ~~9.35 \\
Third part & 14.75 \\
\hline
\end{tabular}
\caption{Average sentiment intensity per trailer section, when
  trailers are divided into three even parts (development set).}
  \vspace{-1em}
\label{tab:sentiment_per_section}
\end{table}

\section{Additional Results}
 Table~\ref{tab:TP_ablation} presents an ablation study highlighting
 the contribution of different modalities (i.e.,~text, video, audio)
 to TP identification. Previous work \cite{papalampidi2020movie} has
 shown that although textual information from \emph{screenplays} is
 very rich, audiovisual cues are also helpful in creating meaningful
 graphs in latent space, which can improve TP identification
 performance. However, in this work we only consider information from
 the movie video and as a result the main network only has access to
 \emph{subtitles} (i.e.,~dialogue parts) without additional
 information about the characters, their actions, or feelings.
We hypothesize that the contribution of audiovisual information in
this case will be larger for identifying important shots in the
video.

Table~\ref{tab:TP_ablation} compares the full model (main and
auxiliary networks are trained jointly; auxiliary network is also
pretrained on more data) against unimodal variants (i.e., the main
network has access to subtitles, video, or audio). We observe that the
visual modality is most informative for identifying key shots in the
movie, while the textual modality is least informative. Perhaps
unsurprisingly, combining all modalities together provides the highest
performance on TP identification.

\section{Task Decomposition Analysis}

\vspace{0.5em} \textbf{Narrative Structure in Trailers} According to
screenwriting theory \cite{Hauge:2017}, the five TPs segment movies
into six thematic units, namely, ``Setup'', ``New Situation'',
``Progress'', ``Complications and Higher Stakes'', ``Final Push'', and
``Aftermath''. To examine which parts of the movie are most likely to
feature in trailers, we compute the distribution of shots per thematic
unit in \emph{gold} trailers (using the extended development set of
TRIPOD). As shown in Figure~\ref{fig:tps_distribution}, trailers on
average contain shots from all sections of a movie, even from the last
two, which might reveal the ending. Moreover, most trailer shots
(30.33\%) are selected from the middle of the movie (i.e.,~Progress) as well as from the beginning (i.e.,~16.62\%
and 25.45\% for ``Setup'' and ``New Situation'', respectively). These
empirical observations corroborate industry principles for trailer
creation.\footnote{\tiny{https://archive.nytimes.com/www.nytimes.com/interactive/2013/02/19/movies/awardsseason/oscar-trailers.html}}

Next, we examine the extent to which the different types of key events
denoted by TPs are present in
trailers. Table~\ref{tab:trailers_vs_trailers} shows the proportion of
trailers (development set) that include at least one shot per TP. As
can be seen, more than half of the trailers (i.e., 52.63\% and
55.26\%) include shots related to the first two TPs, whereas only
34.21\% of trailers contain information about the two final
ones. This is not unexpected,  the first TPs are introductory to
the story and hence more important for making trailers, whereas the
last two may contain spoilers and are often avoided.


\paragraph{Sentiment in Trailers} Empirical rules for
making
trailers\footnote{\tiny{https://www.derek-lieu.com/blog/2017/9/10/the-matrix-is-a-trailer-editors-dream}}
recommend that a trailer should start with shots of medium intensity
to captivate viewers, followed by low intensity in order to deliver
key information about the movie, and finally build up tension until it
reaches a climax. In our work, we approximate intensity flow with
sentiment, and in this section present an empirical analysis of
sentiment flow in real trailers (from our development set).


Specifically, each shot in the trailer is assigned a sentiment
intensity score (regardless of positive or negative polarity).  We
then segment trailers into three equal sections and compute the
average sentiment intensity per section. As shown in
Table~\ref{tab:sentiment_per_section}, the second section is least
intense, whereas the third one is most intense. Finally, when we
measure sentiment flow from one section to the next, we find that
46.67\% of the trailers follow a V-shape, similar to our sentiment
condition for selecting sequences of shots with \graphtrailer.


\begin{table}[t]
\centering
\small
\begin{tabular}{p{8cm}}
\hline
\multicolumn{1}{c}{Comedy/Romance} \\ \hline
'50 First Dates', 'Anger Management', 'As Good As It Gets', 'Bridget Jones Diary', 'Meet the Parents', 'Runaway Bride', 'Sex and The City', 'The Internship', 'Two Weeks Notice', 'Valentine's Day' \\\hline
\multicolumn{1}{c}{Drama/Other} \\\hline
'A Beautiful Mind', 'Cinderella Man', 'Eat Pray Love', 'Forrest Gump',
'Mona Lisa Smile', 'The Curious Case of Benjamin Button', 'The Great
Gatsby', 'The Pursuit of Happyness', 'Wall Street Money Never Sleeps'
\\ \hline
\multicolumn{1}{c}{Action} \\\hline
'Almost Famous', 'American Sniper', 'Australia', 'Cinderella',
'Inception', 'Interstellar', 'Mr and Mrs Smith', 'Night at the Museum:
Battle of the Smithsonian', 'Now You See Me', 'The Da Vinci Code',
'Watchmen' \\ \hline
\multicolumn{1}{c}{Thriller/Mystery} \\ \hline
'Contagion', 'Flightplan', 'Gone Girl', 'Hannibal', 'Outbreak', 'Signs', 'Silent Hills', 'The Insider', 'The Interpret', 'Don't Say A Word' \\
\hline
\end{tabular}
\caption{Movies in the test set and their genres.}
\label{tab:movies_per_genre}
\end{table}

\begin{table}[t]
\centering
\small
\begin{tabular}{L{10em}c}
\hline
 &  Accuracy (\%) \\ \hline
Comedy/Romance & 21.5 \\
Drama/Other & 18.0 \\
Action & 10.5 \\
Thriller/Mystery & 16.4 \\
\hline
\end{tabular}
\caption{\graphtrailer's accuracy for different genres (test set).}
\label{tab:accuracy_per_genre}
\end{table}

\begin{table}[t]
\centering
\small
\begin{tabular}{p{8cm}}
\hline
\multicolumn{1}{c}{Best Trailers} \\\hline
'Contagion', 'Forrest Gump', 'Meet the Parents', 'Mona Lisa Smile', 'Two Weeks Notice' \\\hline
\multicolumn{1}{c}{Worst Trailers} \\ \hline
'Gone Girl', 'Inception', 'Night at the Museum: Battle of the Smithsonian', 'Outbreak', 'Silent Hills' \\
\hline
\end{tabular}
\caption{Movies with best/worst trailers (test set) created
  automatically with \graphtrailer.}
\label{tab:best_worst_movies}
\end{table}

\section{Trailer Creation for Different Genres}

We next break down the performance of our algorithm according to movie
genre.  We first classify the movies in our test set into four broad
genre categories similarly to~\cite{papalampidi2020movie} (see
Table~\ref{tab:movies_per_genre}). Table~\ref{tab:accuracy_per_genre}
reports \graphtrailer's accuracy on different genres (Commedy/Romance,
Drama/Other, Action, and Thriller/Mystery).  Overall, we observe that
the algorithm is more accurate for comedies, romantic movies, and
drama, while it does poorly on action movies. We hypothesize that
action movies consist of short and intense shots in comparison to
comedies and drama and hence it is more challenging to select shots
which are both meaningful out-of-context and exciting for the
``Action'' genre. Finally, the movies with the best and worst
(automatically created) trailers are shown in
Table~\ref{tab:best_worst_movies}. Worst trailers are exclusively
action movies and thrillers.

\begin{table}[t]
\centering
\small
\begin{tabular}{L{10em}c} \hline
 &  Node Connectivity \\ \hline
Comedy/Romance & 0.175 \\
Drama/Other & 0.182 \\
Action & 0.112 \\
Thriller/Mystery & 0.100 \\
\hline
\end{tabular}
\caption{Average node connectivity of trailer shots with respect to
  TP shots in the learned movie graph (development  set). This is a
  measure of how relevant trailer shots are to the movie's storyline depending on genre.}
\label{tab:node_connectivity_trailer_shots}
\end{table}


\begin{figure*}[t]
    \tiny
    \centering
  \begin{tabular}{@{}c@{}@{}c@{}}
        \tiny
        \centering 
        \includegraphics[width=0.48\textwidth]{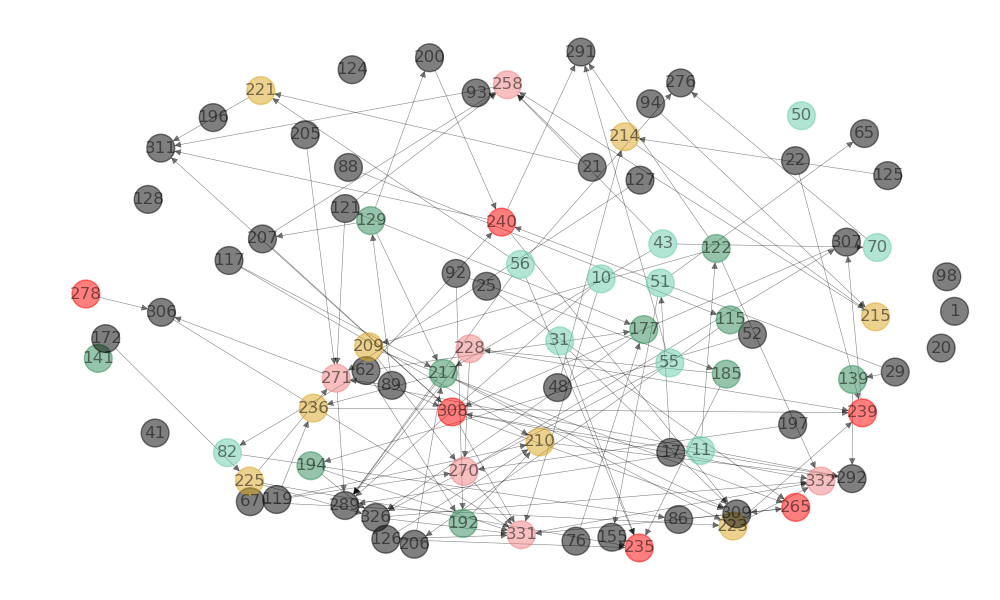} &  \includegraphics[width=0.48\textwidth]{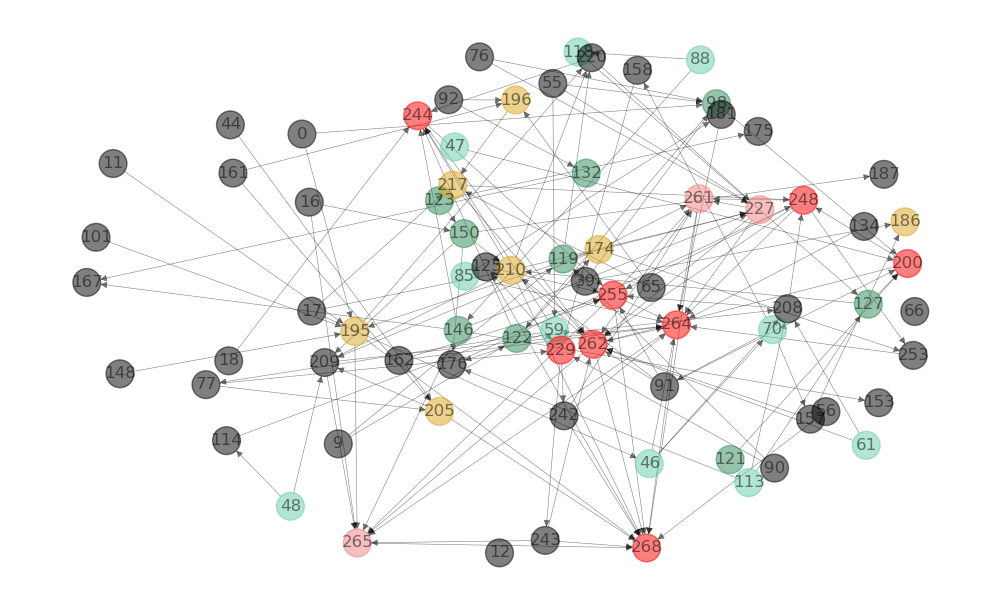} \\
        \small The Wedding Date (Comedy/Romance). & \small Jane Eyre (Drama/Other).
  \end{tabular}
  
  \begin{tabular}{@{}c@{}@{}c@{}}
        \tiny
        \centering 
        \includegraphics[width=0.48\textwidth]{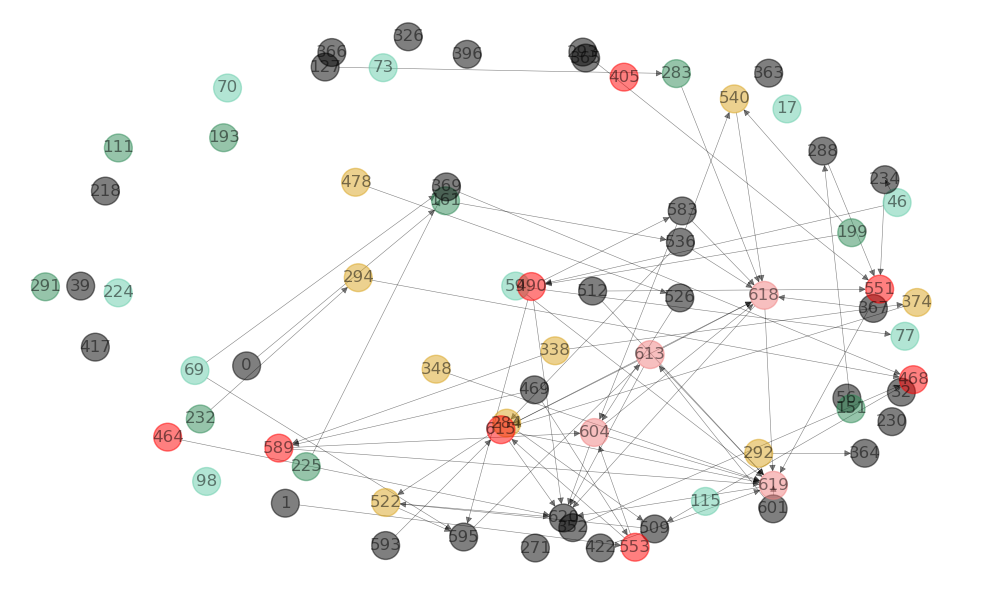} & \includegraphics[width=0.48\textwidth]{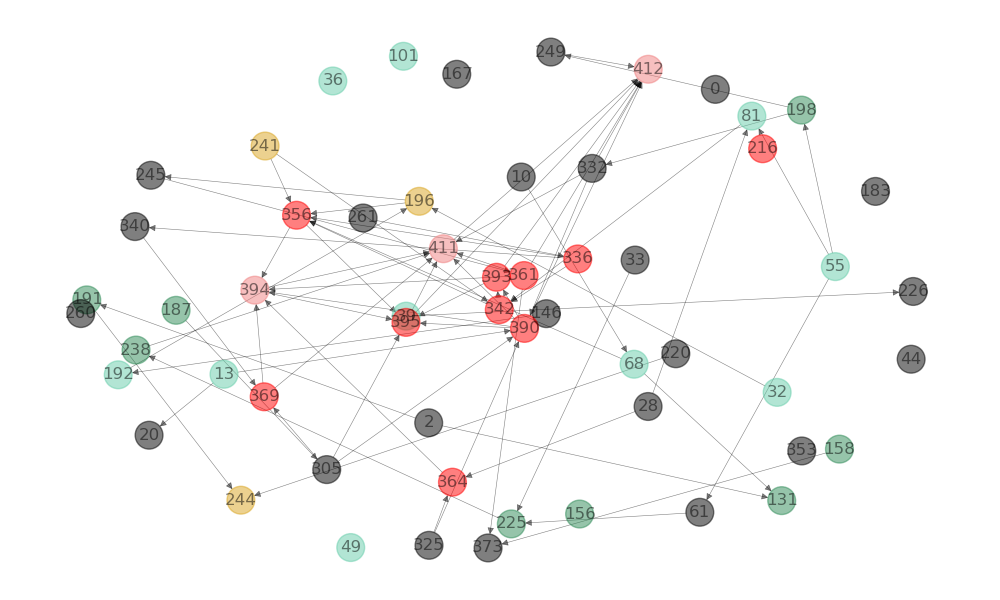} \\
        \small Heat (Action). & \small The Thing (Thriller/Mystery). 
  \end{tabular}
    \caption[ ] {\small Movie graphs (pruned) for four movies (one per
      genre). Graph nodes are turning points (colored) and trailer
      labels (gray nodes). We measure the connectivity between trailer
      and turning point nodes, under the assumption that turning
      points are an approximation of the movie's storyline.}
    \label{fig:pruned_graphs_examples}
\end{figure*}

\begin{figure}[t]
\begin{tabular}{p{8cm}}
\multicolumn{1}{c}{\includegraphics[width=.8\columnwidth]{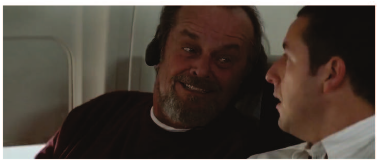}} \\
  \multicolumn{1}{c}{Play status: \textcolor{red}{INCOMPLETE}} \\
  0 seconds out of 61 seconds. (only updates when the video pauses)\\\\
  Please answer the \textbf{following} questions \textbf{immediately after watching} the above
  trailer: \\\hline
  \textbf{1. Information:}  Did you understand what the movie is about based on the trailer?\\
  \radiobutton\hspace{.2cm}Not at all \\
  \radiobutton\hspace{.2cm}Vaguely \\
  \radiobutton\hspace{.2cm}Yes \\ \hline
  \textbf{2. Attractiveness:}  Did you find this trailer aesthetically pleasing? \\
  \radiobutton\hspace{.2cm}No, not at all\\
  \radiobutton\hspace{.2cm}Maybe, not sure \\
  \radiobutton\hspace{.2cm}Yes, I would like to watch the movie \\\hline
  \textbf{3. Spoilers:}   Did the trailer contain any spoilers?\\
  \radiobutton\hspace{.2cm}No, not all \\
  \radiobutton\hspace{.2cm}Maybe, not sure\\
  \radiobutton\hspace{.2cm}Yes, the trailer spoiled the movie \\\hline
  \end{tabular}
    \caption{Questions presented to AMT workers during
      human evaluation: Q1 (information), Q2 (attractiveness), and Q3
      (spoilers). Q3 was only included in the human evaluation study
      of Section~6.2.}
    \label{fig:example_questions}
\end{figure}

Next, we analyze how trailers relate to the movie's storyline and
whether there are genre differences in that regard. We represent a
movie's storyline by the graph learned as part of our TP
identification model. Since these graphs are too large to inspect and
visualize, we follow previous work~\cite{papalampidi2020movie} and
prune them by only keeping the nodes that have been identified as
turning points and those that correspond to (gold) trailer shots. We
use node connectivity to quantify the extent to which trailer shots
relate to the storyline. Specifically, we measure the minimum number
of nodes that need to be removed to separate trailer nodes from TP
nodes into isolated subgraphs.

Table~\ref{tab:node_connectivity_trailer_shots} shows the average node
connectivity of trailer shots per movie genre for the development set.
We also present a few examples of (pruned) graphs with TPs and trailer
shots in Figure~\ref{fig:pruned_graphs_examples}. We observe that
``Comedy/Romance'' and ``Drama/Other'' movies include shots that are
(on average) better connected to the storyline of the movie as
expressed by the identified turning points in comparison to ``Action''
and ``Thriller/Mystery''. This result is intuitive, trailers often
give hints of the plot for romantic comedies and drama, whereas
trailers for action movies and thrillers include intense and exciting
shots (e.g., explosions) that are not necessarily related to the
storyline. Finally, when comparing the results in
Tables~\ref{tab:accuracy_per_genre}
and~\ref{tab:node_connectivity_trailer_shots}, we observe that our
algorithm generally creates better trailers when these are expected to
follow the movie storyline.

\section{Trailer Creation for Different Datasets}

Demonstrating that our approach generalizes to datasets other than
TRIPOD$\bigoplus$ is fraught with difficulty as existing datasets
 are either not publicly available or do not contain all information
 necessary to our task. Specifically:
\begin{itemize}
\item AVA \cite{Gu-2018-CVPR} contains 15-minute video clips instead
  of entire movies, which we need in order to identify trailer shots.
\item MovieNet \cite{huang2020movienet} is a large-scale resource. It
  contains 1,100 movies with trailers and additional metadata
  (e.g.,~photos, plot descriptions) as well as rich manual annotations
  (e.g., of characters, scene boundaries, place and action
  tags). Our method creates trailers over full-length videos which
  MovieNet does not make publicly available, only
  features based on them.
\item VPCD \cite{Brown:ea:2021} contains multi-modal annotations
  (face, body and voice) for all primary and secondary characters from
  a range of diverse TV-shows and movies. Again, only the features are
  released without the corresponding videos, there are no subtitles,
  and the dataset contains episodes from TV series which do not have
  trailers.
\item Condensed Movies \cite{bain2020condensed} does not contain
  entire movies, but rather a few clips per movie collected from
  YouTube. On average, there are only 20 minutes available per movie
  and 36.7\% of the movies do not contain subtitles or closed
  captions. Hence, we cannot use this dataset for our task either.
\end{itemize}

Nevertheless, a key claim in our paper is that narrative structure
which we operationalize by identifying TPs is important for analyzing
movie content and finding trailer moments (i.e.,~shots that could be
potentially included in a trailer).  Our results (see Tables 3 and 6)
further confirm that TPs are integral to the creation of informative
and attractive trailers. Interestingly, the notion of TPs has been
shown to generalize to other datasets and summarization tasks beyond
those exemplified in TRIPOD$\bigoplus$.

For example, Papalampidi et al.~\cite{papalampidi2020screenplay}
illustrate that a TP model trained on the original TRIPOD dataset
\cite{papalampidi2019movie} can be used to identify key scenes in a
dataset consisting of CSI (Crime Scene Investigation) episodes
\cite{frermann2018whodunnit}. They find that narrative structure
representations learned on TRIPOD (which was created for
feature-length films) transfer well across cinematic genres and
computational tasks. Analysis of their video summaries shows that TPs
correlate with information about the crime scene and the victim, the
motive, cause of death, and perpetrator, even though such information
was not explicitly provided at training time.

 Lee et al.~\cite{lee-etal-2021-transformer} propose modifications to
 the TP identification model of Papalampidi et
 al.~\cite{papalampidi2019movie}. They obtain improved performance 
with Transformers instead of LSTMs and by incorporating 
 dialogue information which they claim enhances the model’s
 knowledge between scenes. Their work shows that the original TP model
 can be easily re-configured to accommodate new features (which might
 be useful for different datasets).

 Finally, more recent work focuses on video-to-text
 summarization. Papalampidi and
 Lapata~\cite{papalampidi-lapata-2023-hierarchical3d} create a
 multimodal variant of SummScreen \cite{chen-etal-2022-summscreen}, a
 dialogue summarization dataset consisting of transcripts of TV
 episodes and reference summaries. They introduce the task of
 video-to-text summarization: given a TV episode (video and
 subtitles), generate a recap (in text). To perform this task, they
 first have to select part of the input content since it is too long
 to fit into standard sequence-to-sequence architectures (an episode
 is on average 5.7k tokens long). They employ the TP identification
 model presented here as a content selector. In this case, the model
 is zero-shot transferred to a new domain, slightly different task,
 and a different dataset without any further tuning. Their experiments
 show that content selection based on zero-shot TP identification is
 very competitive against approaches trained on the specific
 summarization task, dataset, and domain.

\section{Human Evaluation Details}

As described in Sections~5.3 and~6.2 of the main paper, we perform
human evaluation in order to judge and compare the quality of trailers
assembled by different methods. We provide an example of the questions
our participants saw in Figure~\ref{fig:example_questions}. AMT
workers first watch the trailers and answer questions like those shown
in Figure~\ref{fig:example_questions}. In the end, they also select
the best and/or worst trailer based on their overall preference. Each
movie was evaluated by five AMT participants. We elicited annotations
for all movies in our test set for both human evaluation studies
(Sections~5.3 and~6.2 of the main paper).

\begin{figure*}[t!]
    \centering
\begin{tabular}{|c|c|} \hline
        \begin{tabular}[c]{c} {{\small Step 1 - Review initial set of shots, metadata:}} \\ {{\small relevance to storyline, sentiment intensity}} \end{tabular} & \begin{tabular}[c]{c} {{\small Optional - Trim shot and select it}} \\ {{\small to move to next step}} \end{tabular}  \\ 
         \includegraphics[width=0.45\textwidth]{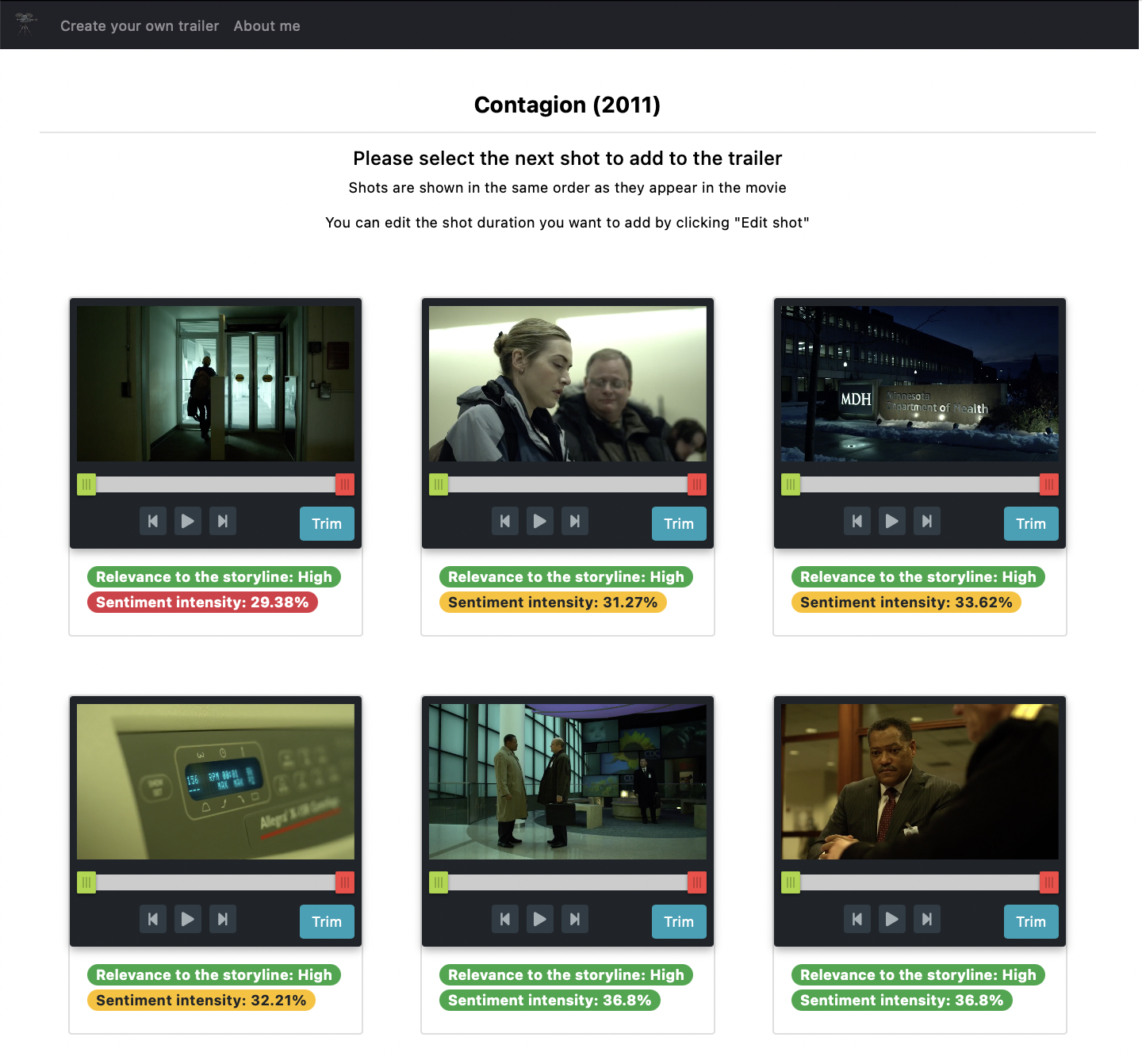} &          
         \includegraphics[width=0.45\textwidth]{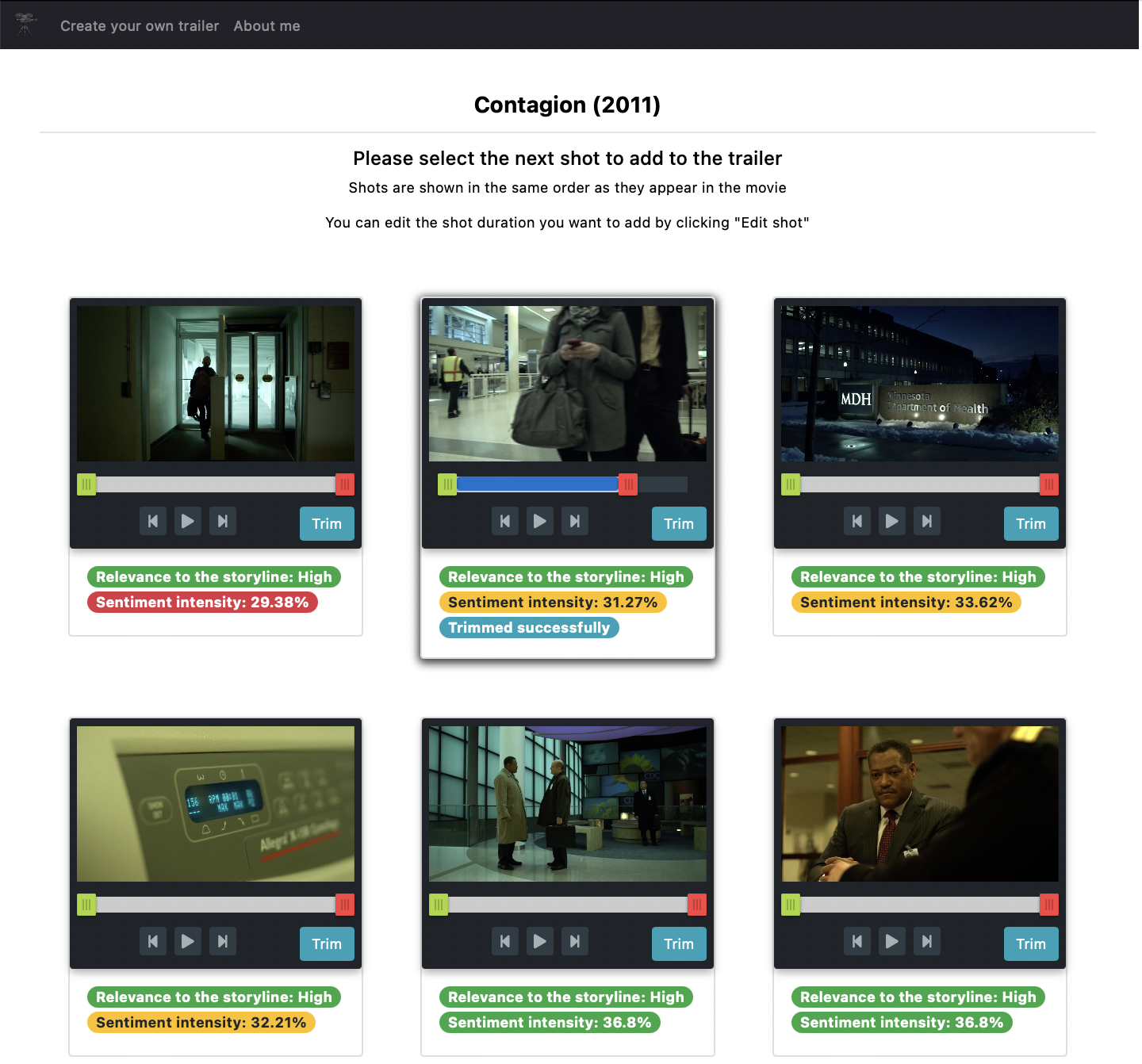} \\ \hline
        
        \begin{tabular}[c]{c} {{\small Step 2 - Review next set of shots, metadata:}} \\ {{\small relevance to storyline, sentiment intensity, transition match}} \end{tabular} & \begin{tabular}[c]{c} {{\small Shots selected so far:}} \\ {{\small keeping track of the path}} \end{tabular}  \\ 
        \includegraphics[width=0.45\textwidth]{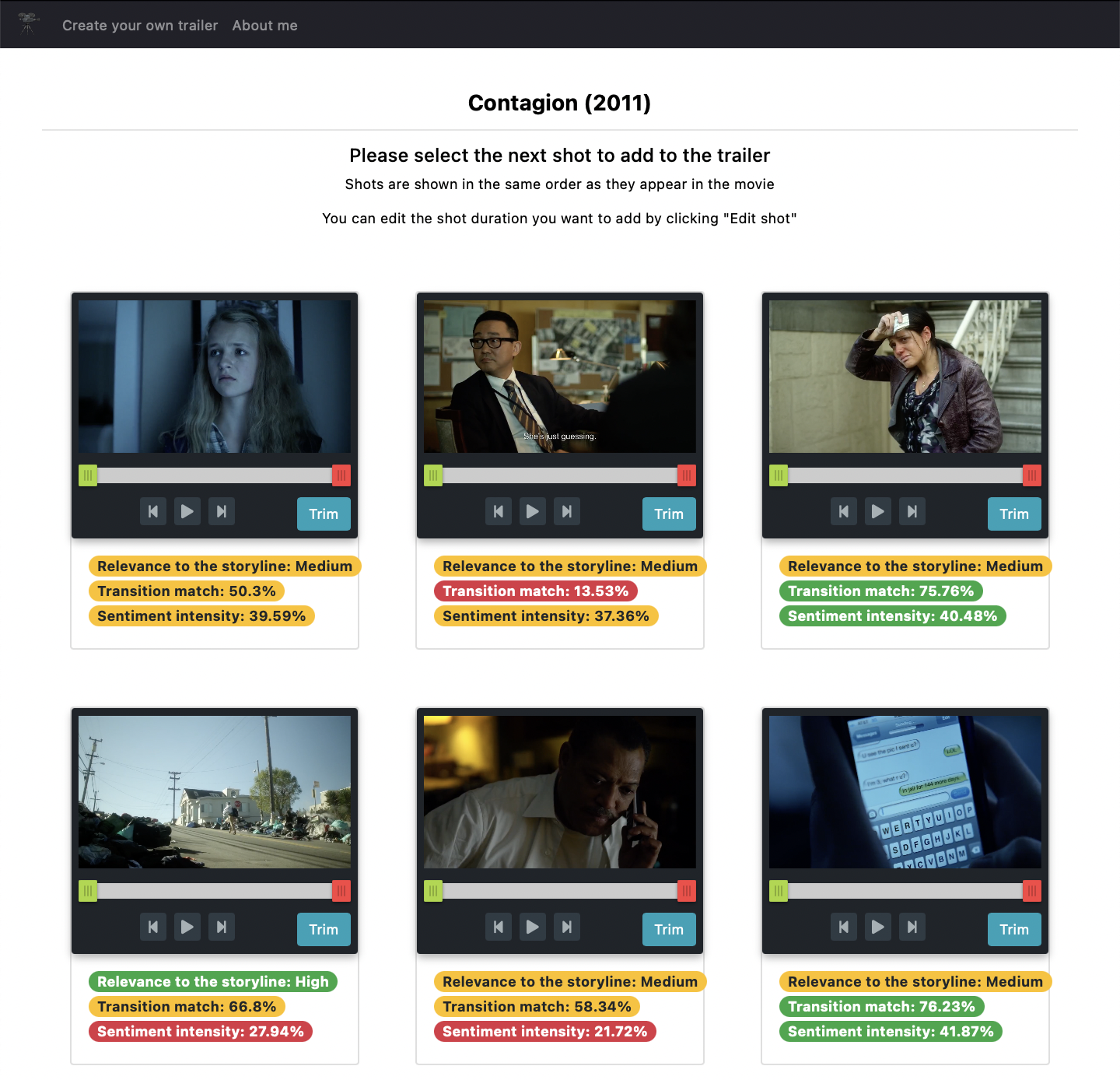} &
         \includegraphics[width=0.42\textwidth]{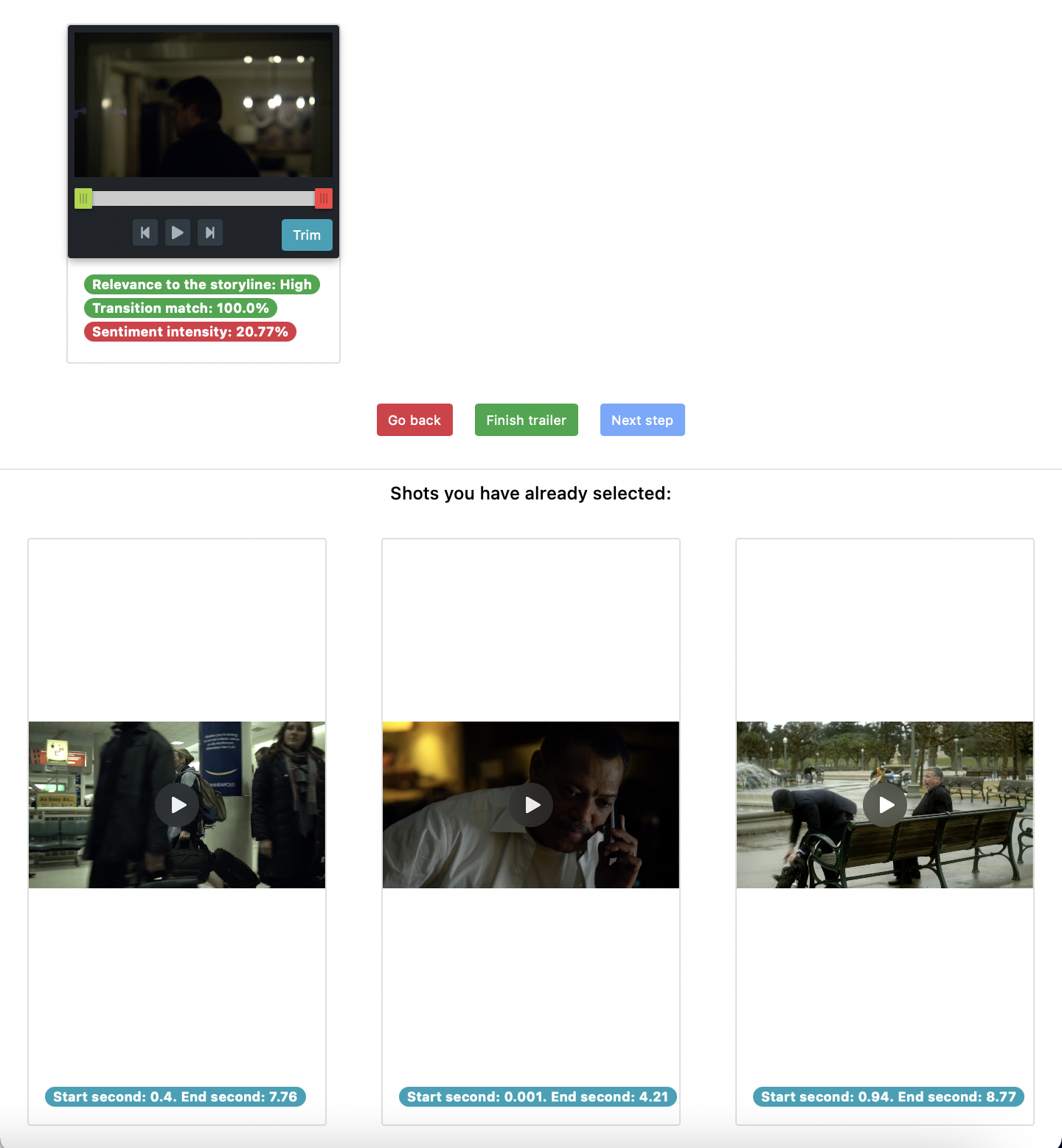} \\ \hline
        
\end{tabular}
 \caption{Interactive trailer creation for the movie ``Contagion''. A
   set of shots are presented to the user who can review them and
   select one (shots can be trimmed to correct for imperfections
   caused by the automatic movie segmentation). A new set of options
   is presented in step~2. The user selects shots interactively, while
   reviewing the trailer being created. They can also go back and
   forth if they are not satisfied with the outcome.  The example
   continues in Figure~\ref{fig:interactive_example_2}.
   \label{fig:interactive_example_1}}
\end{figure*}

\begin{figure*}[t!]
    \centering
\begin{tabular}{|c|c|} \hline
        \begin{tabular}[c]{c} {{\small Final Step - Decide to finish the trailer}} \\ {{\small after selecting a sequence of shots}} \end{tabular} & \begin{tabular}[c]{c} {{\small Assemble and watch final trailer,}} \\ {{\small and (optionally) download the video}} \end{tabular}  \\ 
         \includegraphics[width=0.47\textwidth]{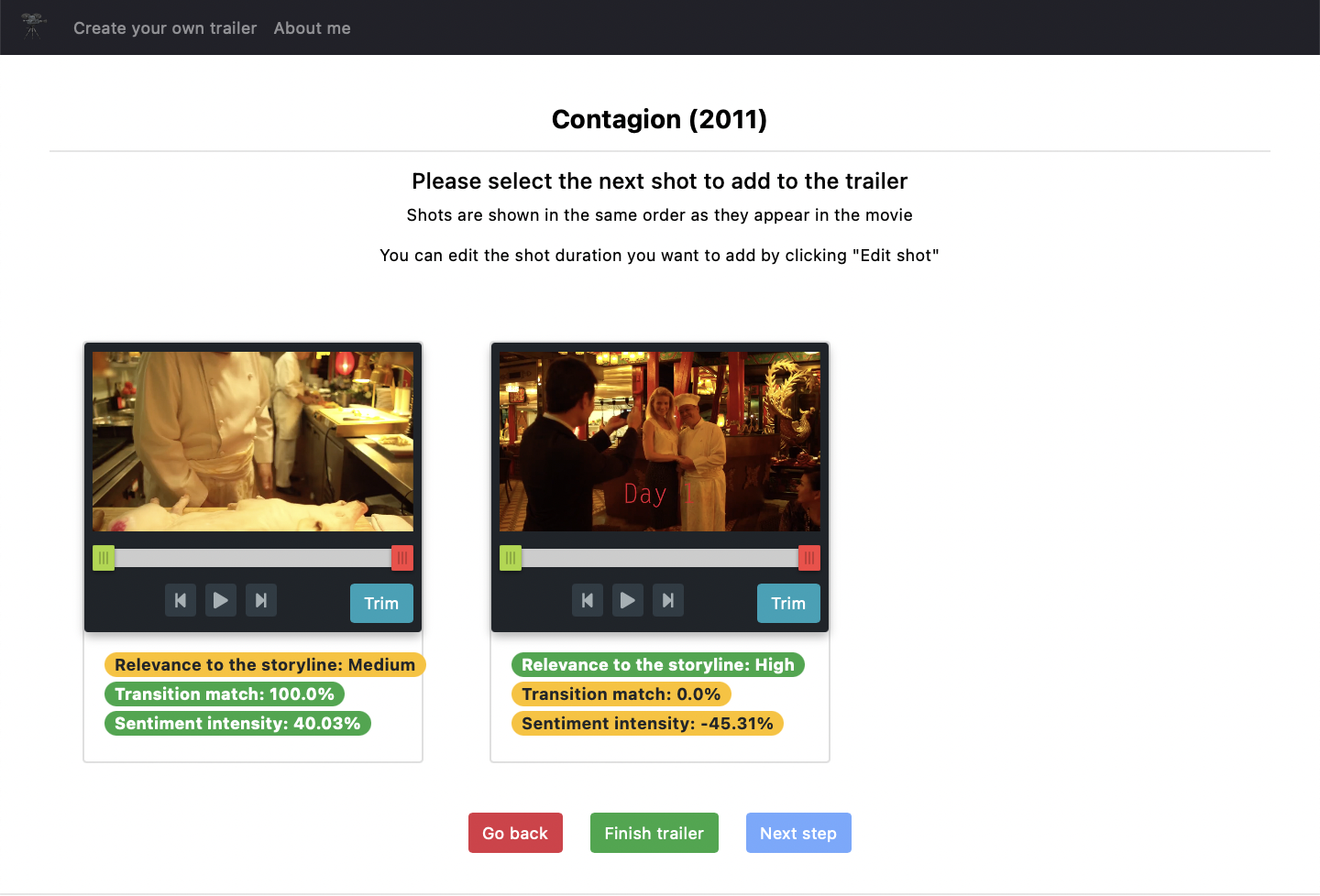} &          
         \includegraphics[width=0.47\textwidth]{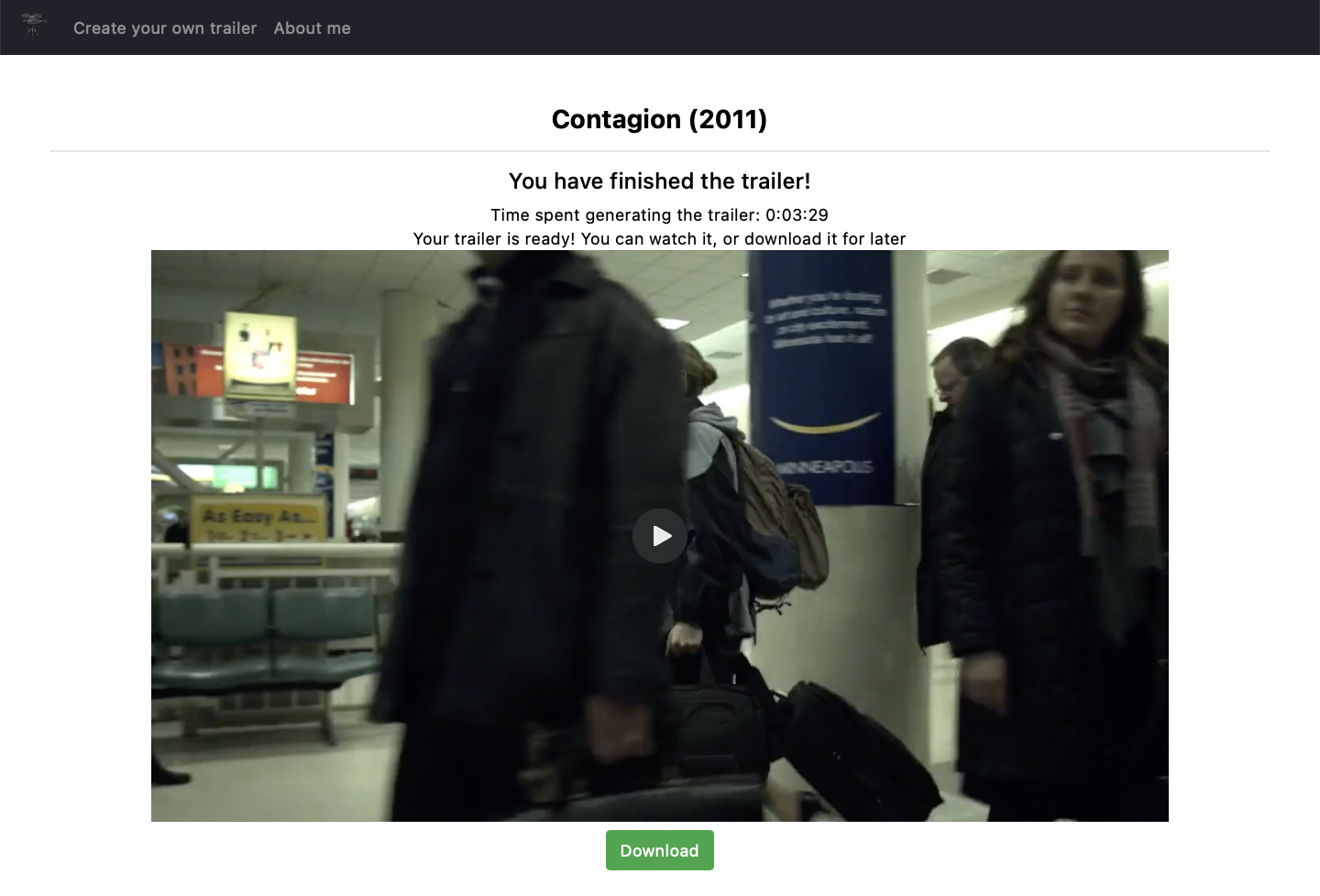} \\ \hline
        
\end{tabular}
 \caption{Finally, the user decides to stop when they are happy
   with their selection. Once the trailer is finished, they can review
   the outcome which is assembled by concatenating the selected shots,
   and optionally download the video.
   \label{fig:interactive_example_2}}
\end{figure*}

\section{Interactive Tool Examples}

In this section we illustrate how the interactive tool for trailer
creation (see also Section~6 of the main paper) works. We consider the
movie ``Contagion'' and explain how a user creates a trailer in
synergy with our algorithm in Figures~\ref{fig:interactive_example_1}
and~\ref{fig:interactive_example_2}.

Initially the user selects the first shot of the trailer from a set of
proposals (Step~1; Figure~\ref{fig:interactive_example_1}). These are
shots are relevant to the storyline as they have all been identified
as TP1. The user can review them alongside metadata (indicated by
green, yellow, and red bars) and decide which one is best for
inclusion in the trailer.  The selected shot can be optionally trimmed
to correct for imperfections caused by the automatic segmentation (see
Trim button in Figure~\ref{fig:interactive_example_1}).  Next, a new
set of options is displayed; the new shots are the  immediate
neighborhood of the  previously selected shot in the
movie graph (see Section~6.1 and Figure~4 of the main paper that
provides an example of how the algorithm works). The user can again
review all new options alongside their metadata (i.e.,~relevance to
storyline, sentiment intensity, and transition match) and decide on
the most appropriate trailer shot (bottom left part of
Figure~\ref{fig:interactive_example_1}).

The user selects shots iteratively, while progressively monitoring the
trailer being created (bottom right part of
Figure~\ref{fig:interactive_example_1}). It is also possible to go
back and forth during the creation process (see button ``Go back'' in
the bottom right part of Figure~\ref{fig:interactive_example_1} and
left part of Figure~\ref{fig:interactive_example_2}) and undo past
decisions. Once satisfied with the outcome (left part of
Figure~\ref{fig:interactive_example_2}), the user can view the final
trailer (assembled by concatenating the selected shots) and optionally
download the video (right part of
Figure~\ref{fig:interactive_example_2}).


\end{document}